\documentclass[lettersize,journal]{IEEEtran}

\usepackage[utf8]{inputenc} % allow utf-8 input
\usepackage[T1]{fontenc}    % use 8-bit T1 fonts
\usepackage{hyperref}       % hyperlinks
\usepackage{url}            % simple URL typesetting
\usepackage{booktabs}       % professional-quality tables
\usepackage{amsfonts}       % blackboard math symbols
\usepackage{nicefrac}       % compact symbols for 1/2, etc.
\usepackage{microtype}      % microtypography
\usepackage{adjustbox}
\usepackage{graphicx, amsmath, amssymb, caption, subcaption, multirow, overpic, textpos}
\usepackage{algorithm, algorithmic}
\usepackage[table,xcdraw]{xcolor}
\begin{document}

\title{Rethinking Multiple Instance Learning for Whole Slide Image Classification: A Good Instance Classifier is All You Need}
%Arbitrary-Shaped Text Detection by Progressive Contour Transformer
%multi-stage contour-based scene text detector with contour transformer

\author{Linhao~Qu,
        Yingfan~Ma,
        Xiaoyuan~Luo,
        Qinhao~Guo,
        Manning~Wang,
        and~Zhijian~Song
\thanks{Manuscript received August, 2023. This work was supported by National Natural Science Foundation of China under Grant 82072021. (Corresponding author: Manning~Wang and Zhijian~Song.)}%
\thanks{Linhao Qu, Yingfan Ma, Xiaoyuan Luo, Manning Wang and Zhijian Song are with Digital Medical Research Center, School of Basic Medical Science, Fudan University, Shanghai 200032, China. They are also with Shanghai Key Lab of Medical Image Computing and Computer Assisted Intervention, Shanghai 200032, China. (e-mail: \{lhqu20, mnwang, zjsong\}@fudan.edu.cn). Qinhao Guo is with Department of Gynecologic Oncology, Fudan University Shanghai Cancer Center, 270 Dong-An Road, Shanghai 200032, China. (e-mail: 18111230017@fudan.edu.cn). }
}

\markboth{IEEE Transactions on Circuits and Systems for Video Technology,~Vol.~X, No.~X, X}%
{Shell \MakeLowercase{\textit{et al.}}: Bare Demo of IEEEtran.cls for IEEE Journals}

\maketitle

\begin{abstract}
    Weakly supervised whole slide image classification is usually formulated as a multiple instance learning (MIL) problem, where each slide is treated as a bag, and the patches 
    cut out of it are treated as instances. Existing methods either train an instance classifier through pseudo-labeling or aggregate instance features into a bag feature through 
    attention mechanisms and then train a bag classifier, where the attention scores can be used for instance-level classification. However, the pseudo instance labels constructed 
    by the former usually contain a lot of noise, and the attention scores constructed by the latter are not accurate enough, both of which affect their performance. 
    In this paper, we propose an instance-level MIL framework based on contrastive learning and prototype learning to effectively accomplish both instance classification and bag 
    classification tasks. To this end, we propose an instance-level weakly supervised contrastive learning algorithm for the first time under the MIL setting to effectively learn 
    instance feature representation. We also propose an accurate pseudo label generation method through prototype learning. We then develop a joint training strategy for weakly 
    supervised contrastive learning, prototype learning, and instance classifier training. Extensive experiments and visualizations on four datasets demonstrate the powerful 
    performance of our method. Codes are available at \url{https://github.com/miccaiif/INS}.
\end{abstract}


\section{Introduction}

\begin{figure}[t!]
   \begin{center}
   \includegraphics[width=1\linewidth]{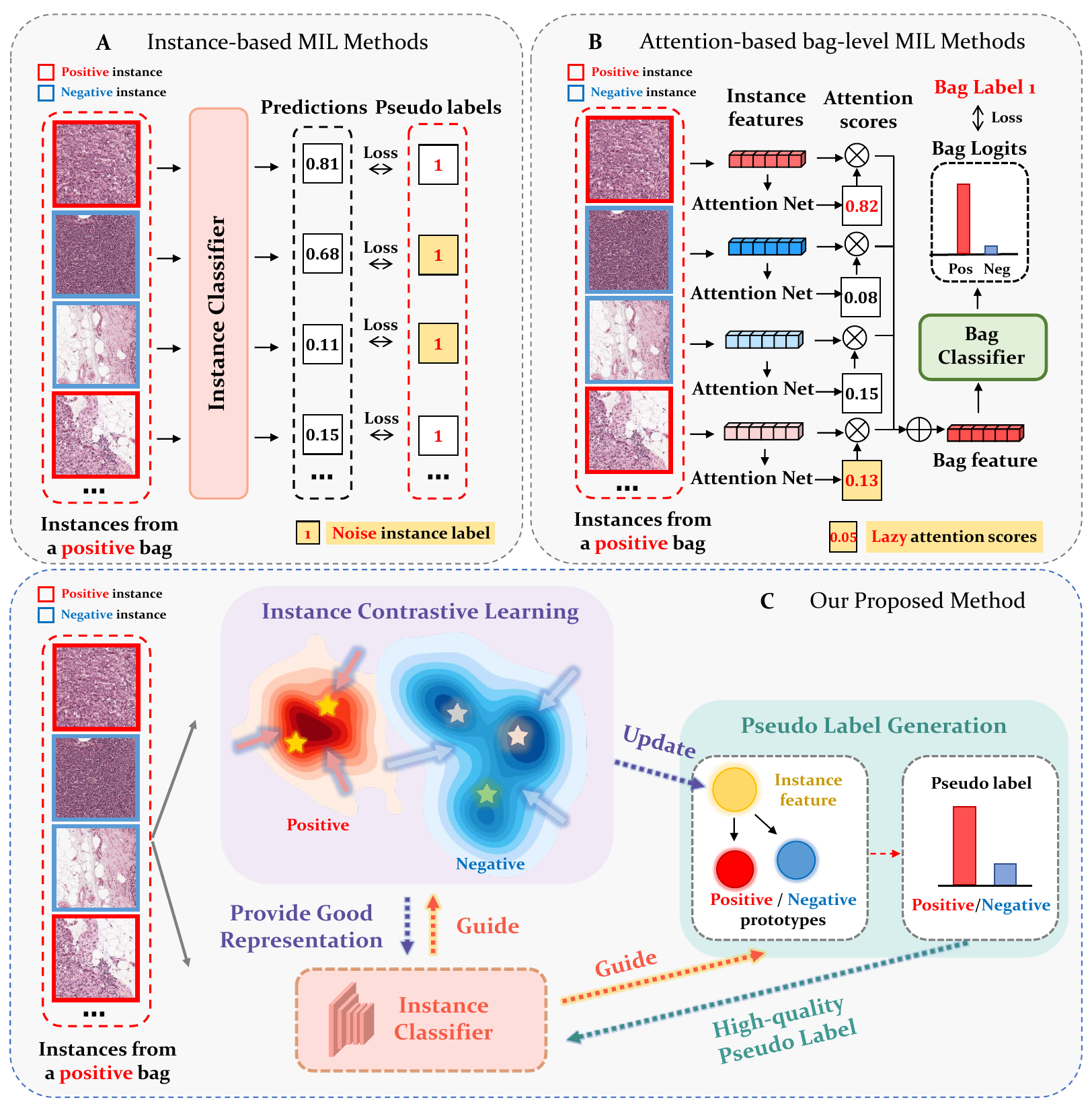}
   \end{center}
   % \vspace{-0.8em}
   \caption{Motivation of our method. (A). Existing instance-based MIL methods typically assign a bag's label to its instances as pseudo labels, resulting in a large number of noises in instance pseudo labels. 
      (B). The loss function of bag-based MIL methods is defined at the bag level, which often only finds the most easily identifiable positive instances and ignores other more difficult ones. 
      (C). In contrast to these methods, we propose an effective instance-based MIL framework based on contrastive learning and prototype learning and a joint training strategy. Our main goal is to develop an effective instance classifier using instance-level weakly supervised contrastive learning and pseudo-label generation. Contrastive learning aims to improve instance representations to better distinguish between negative and positive instances in the feature space. High-quality pseudo-labels are then generated for each instance based on representative feature vectors. Both feature learning and pseudo-label generation are guided by the instance classifier, enhancing its capabilities.}
   \label{figure1}
   % \vspace{-2em}
\end{figure}

\IEEEPARstart{T}{he} deep learning-based whole slide image (WSI) processing technology is expected to greatly promote the automation of pathological image diagnosis and analysis \cite{24,25,36,38,39,100,110}. However, 
WSIs are quite different from natural images in size, and the size can range from 100 million to 10 billion pixels, which makes it impossible to directly utilize deep learning models developed for natural 
images to WSIs. It is a common approach to divide WSI into many non-overlapping small patches for processing, but providing fine-grained annotations for these patches is very expensive 
(a WSI can typically produce tens of thousands of patches), making patch-based supervised methods infeasible \cite{26,27,34,35,37}. Therefore, weakly supervised learning approaches based on 
Multiple Instance Learning (MIL) have become the mainstream in this field. In the MIL setting, each WSI is regarded as a bag, and the small patches cut out of it are regarded as instances 
of the bag. For a positive bag, there is at least one positive instance, while for a negative bag, all instances are negative. In clinical applications, there are two main tasks for WSI 
classification: bag-level classification, which accurately predicts the class of a whole slide, and instance-level classification, which accurately identifies positive instances \cite{3,18}.

Currently, MIL methods for WSI classification can be divided into instance-based methods \cite{1,2,3,28} and bag-based methods \cite{8,9,11,12,13,14,15,30,37,44,42}. Instance-based methods typically 
train an instance classifier with pseudo labels and then aggregate the predictions of instances to obtain the bag-level prediction. The main problem of this 
approach is that \textbf{instance pseudo labels contain a lot of noise}. Since the true label of each instance is unknown, the quality of the pseudo labels assigned to all instances is a key factor that 
determines the performance of this type of methods. However, existing studies usually assign instance pseudo labels by inheriting the label of the bag they belong to, which leads to a large amount 
of noise in the pseudo labels and greatly limits their performance. Figure \ref{figure1} A shows the typical training process and issues of instance-based methods.

Bag-based methods first use an instance-level feature extractor to extract features for each instance in a bag, and then aggregate these features to obtain a bag-level feature, 
which is used to train a bag classifier. Most recent bag-based methods utilize attention mechanisms to aggregate instance features and they introduce an independent scoring module 
to generate learnable attention weights for each instance feature, which can be used to realize instance-level classification. Although this type of method overcomes the problem of 
noisy labels in instance-based methods, it also has the following issues: \textbf{1) Low performance in instance-level classification.} We find that the difficulty of identifying different 
positive instances is different in the same positive bag (\textit{e.g.}, instances with larger tumor areas are easier to be identified than those with smaller tumor areas). Attention-based 
methods define losses at the bag level, which often leads to the result that only the most easily identifiable positive instances are found through the high attention scores while other more difficult ones are missed \cite{12,18}. In other words, the network only assigns higher attention weights to the most easily recognizable positive instances to achieve correct bag classification, without the 
need to find all positive instances, which greatly limits its instance-level classification performance. \textbf{2) Bag-level classification performance is not robust.} As mentioned earlier, 
bag-level classification relies heavily on the attention scores assigned by the scoring network to each instance. When these attention scores are inaccurate, the performance of the 
bag classifier will also be affected. A typical example is the bias that occurs in classifying bags with a large number of difficult positive instances while very few easy positive 
instances. Figure \ref{figure1} B shows the typical training process and issues of bag-based methods.

In the history of deep learning-based MIL research, although instance-based methods were first proposed, their reliance on instance pseudo labels, which are difficult to obtain, 
has led to a bottleneck in their performance. In contrast, an increasing number of researchers \cite{13,15} have focused on using stronger attention-based methods at the bag level to 
provide more accurate attention scores. However, intuitively, as long as the loss function is defined at the bag level, attention-based scoring methods will inevitably exhibit 
"laziness" in finding more difficult positive instances \cite{12,18}. Different from the above-mentioned studies, we propose an instance-based MIL framework based on contrastive 
learning and prototype learning, called INS. Figure \ref{figure1} C illustrates the main idea and basic components of INS. Our main objective is to directly train an efficient instance 
classifier at the fine-grained instance level, which needs to fulfill two requirements: first, obtaining a good instance-level feature representation, and second, assigning 
an accurate pseudo label to each instance. To this end, we propose instance-level weakly supervised contrastive learning (IWSCL) for the first time in the MIL setting to learn 
good instance representations, better separating negative and positive instances in the feature space. We also propose the Prototype-based Pseudo Label Generation (PPLG) strategy, 
which generates high-quality pseudo labels for each instance by maintaining two representative feature vectors as prototypes, one for negative instances and the other for positive 
instances. We further develop a joint training strategy for IWSCL, PPLG and the instance classifier. Overall, IWSCL and PPLG are completed under the guidance of the instance 
classifier's predictions. At the same time, the good feature representations from IWSCL and the accurate pseudo labels from PPLG can then further improve the instance classifier. 
More importantly, we efficiently utilize the true negative instances from negative bags in the training set to guide all the instance classifier, the IWSCL and the PPLG, ensuring 
that INS iterates towards the right direction. After training the instance classifier, due to its strong instance classification performance, we can complete accurate bag 
classification using simple mean pooling.

We comprehensively evaluated the performance of INS on six tasks, using a simulated CIFAR10 dataset and three real-world datasets containing breast cancer, 
lung cancer, and cervical cancer. Extensive experimental results demonstrate that INS achieved better performance in both instance and bag 
classification than state-of-the-art methods. More importantly, our experiments not only include the tasks that human doctors can directly judge from H$\&$E-stained 
slides, like tumor diagnosis and tumor subtyping, but also the tasks that human doctors cannot directly make decisions from HE slides, including predicting lymph 
node metastasis from primary lesion, patient prognosis, and prediction of immunohistochemical markers. Given the strong instance-level classification ability of INS, 
we can use it for explainable research and new knowledge discovery in these difficult clinical tasks. In the task of predicting lymph node metastasis from primary lesion 
of cervical cancer, we use INS to classify high- and low-risk instances, thereby identifying the "Micropapillae" pathological pattern that indicates high risk of lymph node metastasis.

\textbf{The main contributions of this paper are as follows:}

\noindent$\bullet $ We propose INS, an instance-based MIL framework that combines contrastive learning and prototype learning. This framework serves as an efficient instance classifier, capable of effectively addressing instance-level and bag-level classification tasks at the finest-grained instance level.

\noindent$\bullet $ We propose instance-level weakly supervised contrastive learning (IWSCL) for the first time in the MIL setting to learn good feature representations for each instance. We also propose the Prototype-based Pseudo Label Generation (PPLG) strategy, which generates high-quality pseudo labels for each instance through prototype learning. We further propose a joint training strategy for IWSCL, PPLG, and the instance classifier.

\noindent$\bullet $ We comprehensively evaluated the performance of INS on six tasks of four datasets. Extensive experiments and visualization results demonstrate that INS achieves the best performance of instance and bag classification.

%-------------------------------------------------------------------------
\section{Related Work}
%-------------------------------------------------------------------------
\subsection{Instance-based MIL Methods}
Instance-based methods train an instance classifier by assigning pseudo labels to each instance, and bag classification is achieved by aggregating the prediction of all instances in a bag. 
Early methods \cite{4,5,6,7} typically assign a bag's label to all its instances, leading to a large number of noisy labels in positive bags. Some recent methods \cite{1,2,3} select key instances and only use them 
for training, thus reducing the impact of noise to some extent. In this paper, we present a strong instance classifier and we believe that a good instance classifier requires both good instance-level 
representation learning and accurate pseudo instance labels. To fulfill this goal, we propose for the first time the instance-level weakly supervised contrastive learning under the MIL setting, 
which achieves efficient instance feature representation. We also propose a prototype learning-based strategy to generate high-quality pseudo labels.
%-------------------------------------------------------------------------
\subsection{Bag-based MIL Methods}
Bag-based methods first extract instance features and then aggregate these features in a bag to generate bag features for training a bag classifier. 
Attention-based methods \cite{8,9,11,12,13,14,37,111} are the mainstream of this paradigm, which typically use an independent scoring network for each instance feature to produce 
learnable attention weights, which can also be used to generate instance predictions. The main problem of these methods is that they cannot accurately identify difficult 
positive instances, resulting in limited instance and bag classification performance. Recently, some studies \cite{12,18,44} have added instance-level classification loss to 
bag-level losses, but the pseudo labels assigned to instances are still noisy.
Some methods have also been proposed to accomplish WSI classification using reinforcement learning \cite{92}, graph learning \cite{94,99}, and bayesian learning \cite{93}. 
However, none of them can effectively fulfill the instance classification task. In contrast, we directly start from the finest instance-level and use weakly supervised contrastive 
learning and prototype learning to complete instance feature learning and pseudo-label updating, thereby addressing both instance and bag classification.
%-------------------------------------------------------------------------

\begin{figure}[htbp]
\begin{center}

\includegraphics[width=\linewidth]{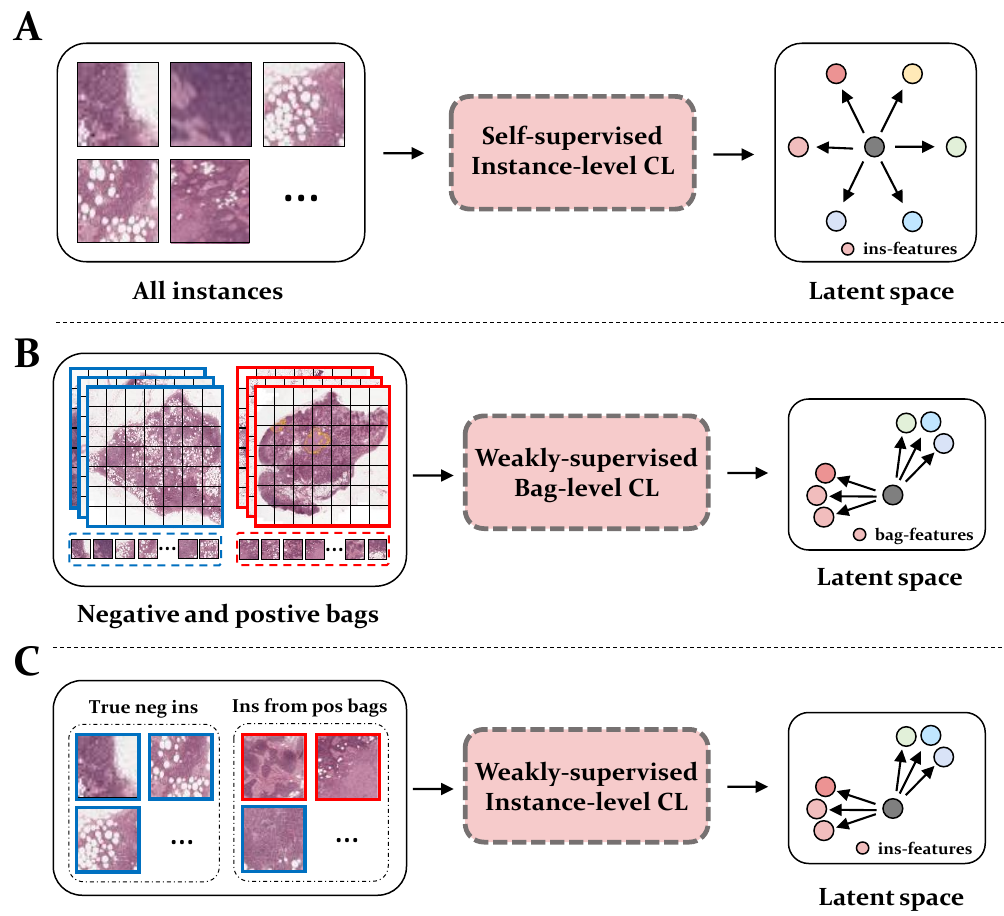}
\end{center}
%  \vspace{-1em}
   \caption{Comparison of our instance-level weakly supervised contrastive learning (IWSCL)} with existing contrastive learning methods. The red box and blue box represent positive and negative instances, respectively. 
   "Ins" is the abbreviation for instances, "pos" for positive, "neg" for negative, and "CL" for contrastive learning. A. Self-supervised instance-level contrastive 
   learning separates instance features as much as possible, regardless of whether they are positive or negative. B. Weakly supervised contrastive learning at the bag level \cite{42} does not take instance features into account. 
   C. The proposed IWSCL tries to separate positive and negative instances directly in the feature space.
    %   \vspace{-1.5em}
\label{figure2}
\end{figure}
%-------------------------------------------------------------------------
  %\vspace{-0.5em}
\subsection{Contrastive Learning for WSI Classification}
\label{CL}
Existing methods \cite{13,14,15,18,40,41,43} usually first use WSI patches to pretrain an instance-level feature extractor through self-supervised learning and then perform model training 
using the extracted features. Most of them \cite{13,14,18,41,43} use contrastive self-supervised learning methods \cite{31,46,47,105,106,107,108} to extract instance features, but this process is completely 
unsupervised and it can only attempt to separate all instances as much as possible instead of effectively separating positive and negative instances. In the latest research, Wang et al. \cite{42} 
proposed a feature-based contrastive learning method at the bag-level, but it still cannot perform effective instance classification. In contrast, we for the first time propose instance-level 
weakly supervised contrastive learning (IWSCL) under the MIL setting, which effectively separates negative and positive instance features. Figure \ref{figure2} shows the differences between existing 
contrastive learning methods and our proposed IWSCL.
%-------------------------------------------------------------------------

\subsection{Prototype Learning for WSI Classification}
\label{PL}
Prototype learning, derived from Nearest Mean Classifiers, aims to provide a concise representation for instances. Recent studies demonstrate the potential of using representations or prototypes 
for classification, with variations in construction and utilization \cite{101,102,103,104}. ProtoMIL \cite{109} initially extracts features from each patch using a convolutional layer, then builds negative and positive prototypes by comparing them through similarity calculations, and finally aggregates them using attention scores. PMIL \cite{90} employs unsupervised clustering to construct prototypes, enhancing bag features by assessing instance similarity within bags. 
TPMIL \cite{91} creates learnable prototype vectors, utilizing attention scores as soft pseudo-labels to assign instances. However, PMIL's non-learnable prototypes focus on improving bag features, posing 
challenges for fine-grained instance classification. ProtoMIL and TPMIL heavily relies on attention scores, which often fail to accurately identify challenging positive instances. Additionally, their prototype learning 
lacks effective integration of feature-level contrastive learning, resulting in limited performance. In contrast, our proposed PPLG strategy generates high-quality pseudo-labels, utilizing joint training 
with instance contrastive representation learning, prototype learning, and the instance classifier. Our prototype learning also incorporates guidance from the instance classifier and includes true negative 
instances from negative bags in the training set.
%-------------------------------------------------------------------------

\section{Method}
\subsection{Problem Formulation}
Given a dataset $X=\{X_1,X_2,\ldots,X_N\}$ containing $N$ WSIs, and each WSI $X_i$ is divided into non-overlapping patches $\{x_{i,j},j=1,2,\ldots n_i\}$, where $n_i$ denotes the number of patches 
obtained from $X_i$. All the patches from $X_i$ constitute a bag, where each patch is an instance of this bag. The label of the bag $Y_i\in\left\{0,1\right\},\ i=\{1,2,...N\}$, and the labels of 
each instance $\{y_{i,j},j=1,2,\ldots n_i\}$ have the following relationship:
\begin{equation}
   Y_i=\left\{\begin{array}{cc}
      0,& \quad \text { if } \sum_j y_{i, j}=0 \\
      1,& \quad \text { else }
   \end{array}\right.
   \label{eq1}
\end{equation}

This indicates that all instances in negative bags are negative, while in positive bags, there exists at least one positive instance. In the setting of weakly supervised MIL, only the labels of 
bags in the training set are available, while the labels of instances in positive bags are unknown. Our goal is to accurately predict the label of each bag (bag classification) and the label of each 
instance (instance classification) in the test set.

\subsection{Framework Overview}
\begin{figure*}[htbp]
    \begin{center}
    \includegraphics[width=\linewidth]{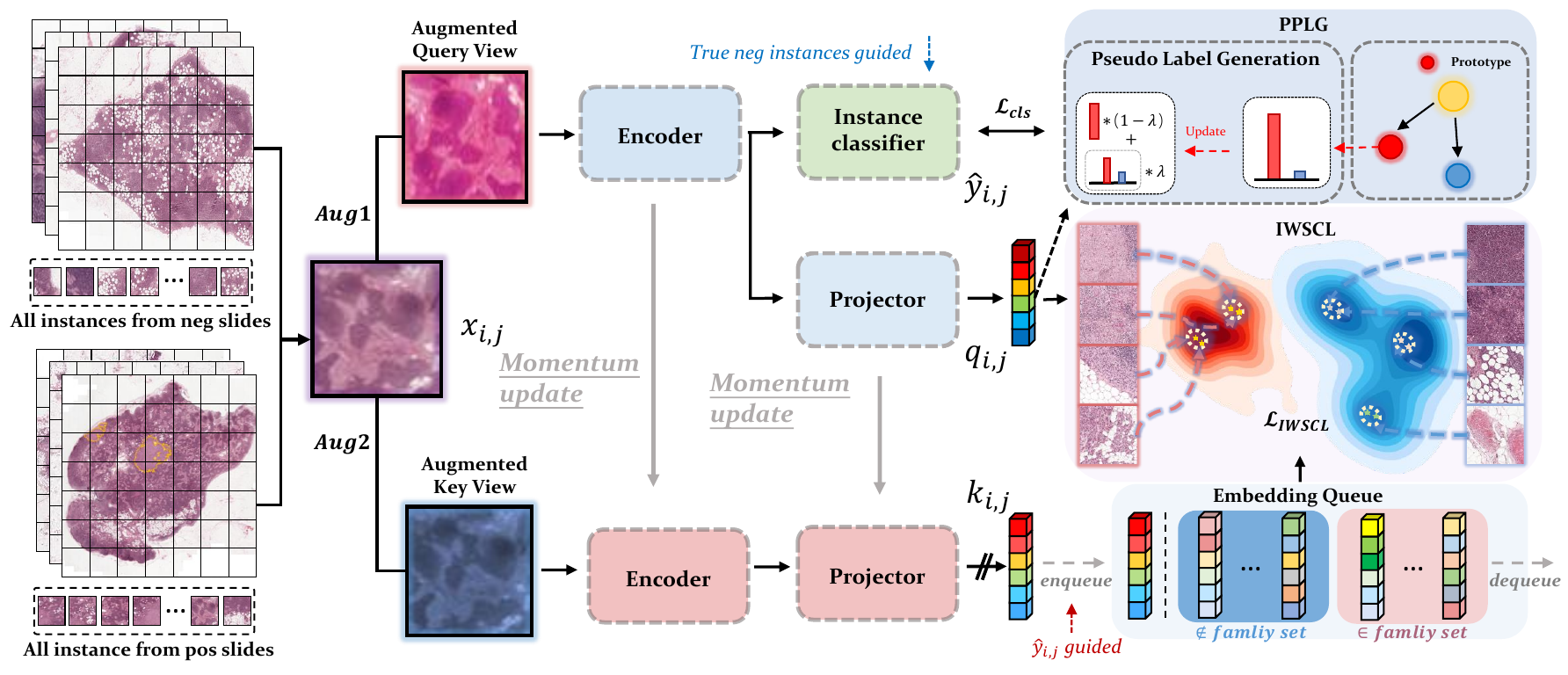}
    %\fbox{\rule{0pt}{2in} \rule{.9\linewidth}{0pt}}
    \end{center}
 %    \vspace{-1.2em}
       \caption{Workflow of the proposed INS framework, where "pos" is the abbreviation for positive, "neg" for negative, "Aug" for augmentation, "IWSCL" for the proposed instance-level 
       weakly supervised contrastive learning, and "PPLG" for the proposed prototype-based pseudo label generation.}
 %    \vspace{-1.5em}
    \label{figure3}
 \end{figure*}

Figure \ref{figure3} presents the overall framework of the proposed INS, which aims to train an efficient instance classifier using instance-level weakly supervised contrastive learning (IWSCL) and 
Prototype-based Pseudo Label Generation (PPLG). We use the true negative instances from negative bags in the training set to guide all the instance classifier, the IWSCL and the 
PPLG, ensuring that INS iterates towards the right direction. IWSCL and PPLG are also guided by the instance classifier, and they also help improve the instance classifier through iterative optimization.

Specifically, in one iteration, we first randomly select an instance $x_{i,j}$ from all instances in the training set, and generate a query view and a key view using two different augmentations. 
In the Query View branch, we input the query view into an Encoder and then feed its output to both the instance classifier and the MLP-based projector to obtain the 
predicted class ${\hat{y}}_{i,j}\in\mathbb{R}^2$ (a one-hot vector indicating negative or positive) and the feature embedding $q_{i,j}\in\mathbb{R}^d$, respectively. 
In the Key View branch, we input the key view into an Encoder and then feed its output to a Projector to obtain the feature embedding $k_{i,j}\in\mathbb{R}^d$, where 
both Encoder and Projector are updated through momentum-based methods from the Query View branch. Inspired by MOCO \cite{46} and Pico \cite{48}, we maintain a large Embedding Queue 
to store the feature embeddings of the Key View branch together with the predicted class labels of the corresponding instances. Then, we use the 
current instance's ${\hat{y}}_{i,j}$,\ $q_{i,j}$, $k_{i,j}$, and the Embedding Queue from the previous iteration to perform instance-level weakly supervised contrastive 
learning (IWSCL). For $q_{i,j}$, we pull closer the instance embeddings in the Embedding Queue that have the same predicted class and push away those with different 
predicted class. In the PPLG module, we maintain two representative feature vectors as prototypes for positive and negative classes during training. We use $q_{i,j}$ and 
the prototype vectors from the previous iteration to generate pseudo labels for $x_{i,j}$. Finally, we update the Embedding Queue using ${\hat{y}}_{i,j}$ and $k_{i,j}$, update 
the prototype vectors using ${\hat{y}}_{i,j}$ and $q_{i,j}$, and train the instance classifier with the generated pseudo label, which completes the current iteration. 
In addition, to prevent bag-level degradation during training, we add a bag-level constraint loss function. The IWSCL module and the PPLG module are presented in Section \ref{sec33} and Section \ref{sec34}, respectively. The bag constraint and total loss are given in Section \ref{sec35}.

\subsection{Instance-level Weakly Supervised Contrastive Learning}
\label{sec33}
In contrastive learning, the most important step is to construct positive and negative sample sets, and then learn robust feature representations by pulling positive samples closer and pushing 
negative samples farther in the feature space \cite{31,46,47}. To distinguish from the positive and negative instances in the MIL setting, 
we use family/non-family sample sets to represent the positive/negative sample sets in contrastive learning, respectively.

In traditional self-supervised contrastive learning, the standard method for constructing family and non-family sets is to use two augmented views of the same sample as family samples, 
while all other samples are considered as non-family sets. This can only force all samples to be as far away from each other as possible in the feature space, but cannot separate 
positive and negative instances in the MIL setting. In contrast, in the MIL setting, all instances from negative bags in the training set have true negative labels, and they 
naturally belong to the same set. This weak label information can effectively guide the instance-level contrastive learning, which is neglected in existing studies. We 
maintain a large Embedding Queue during training to store the feature embeddings $k_{i,j}$ of a large number of instances and their predicted classes ${\hat{y}}_{i,j}$ by the 
instance classifier. Note that for true negative instances, we no longer save their predicted classes, but directly assign them a definite negative class, \textit{i.e.}, ${\hat{y}}_{i,j}$=0.

\textbf{Family and Non-family Sample Selection.} For the instance $x_{i,j}$ and its embedding $q_{i,j}$, we use the instance classifier's predicted class ${\hat{y}}_{i,j}$ and the 
Embedding Queue to construct its family set $F(q_{i,j})$ and non-family set $F'(q_{i,j})$, and then perform contrastive learning based on $q_{i,j}$. Specifically, $F(q_{i,j})$ comes 
from two parts, of which the first part consists of the embeddings $q_{i,j}$ and $k_{i,j}$ and the second part consists of all embeddings in the Embedding 
Queue whose class label equals ${\hat{y}}_{i,j}$. Embeddings with the other class label in the Embedding Queue form the non-family set $F'(q_{i,j})$.

Mathematically, for a given mini-batch, let all query and key embeddings be denoted as $B_q$ and $B_k$, and the Embedding Queue as $Q$. 
For an instance $(x_{i,j},q_{i,j},{\hat{y}}_{i,j})$, its contrastive embedding pool is defined as:
\begin{equation}
   P\left(q_{i, j}\right)=\left(B_q \cup B_k \cup Q\right) \backslash\left\{q_{i, j}\right\}  \label{eq2}
\end{equation}

In $P(q_{i,j})$, its family set $F(q_{i,j})$ and non-family set $F'(q_{i,j})$ are defined as:
\begin{equation}
   F\left(q_{i, j}\right)=\left\{m \mid m \in P\left(q_{i, j}\right), \hat{y}_m=\hat{y}_{i, j}\right\}  \label{eq3}
\end{equation}
\begin{equation}
   F'\left(q_{i, j}\right)=P\left(q_{i, j}\right) \backslash F\left(q_{i, j}\right)  \label{eq4}
\end{equation}

\textbf{Contrastive Loss.} We construct a contrastive learning loss based on the embedding $q_{i,j}$:
% \begin{small}
\begin{equation}
   \begin{aligned}
      &\mathcal{L}_\mathit{I W S C L}\left(q_{i, j}\right)=\\
      &-\frac{1}{\left|F\left(q_{i, j}\right)\right|} \sum_{{k}_{+} \in F\left(q_{i, j}\right)} \log \frac{\exp \left(q_{i, j}{ }^{\top} {k}_{+} / \tau\right)}{\Sigma_{{k}^{-} \in F^{\prime}\left(q_{i, j}\right)} \exp \left(q_{i, j}{ }^{\top} {k}_{-} / \tau\right)},
   \end{aligned}
   \label{eq5}
\end{equation}
% \end{small}
where ${k}_{+}$ denotes the family sample of the current $q_{i,j}$, ${k}_-$ denotes the non-family sample of the current $q_{i,j}$, and $\tau\geq0$ is the temperature coefficient.

\textbf{Embedding Queue Updating.} At the end of each iteration, the current instance's momentum embedding $k_{i,j}$ and its predicted label ${\hat{y}}_{i,j}$ or true negative label are added to the Embedding Queue, 
and the oldest embedding and its label are dequeued.

\subsection{Prototype-based Pseudo Label Generation}
\label{sec34}
On the basis of obtaining a meaningful feature representation, we assign more accurate pseudo labels to instances by prototype learning. To this end, we maintain 
two representative feature vectors, one for negative instances and the other for positive instances, as prototype vectors ${\mu}_r\in\mathbb{R}^d,{r=0,1}$. 
The generation of pseudo labels and the updating process of prototypes are also guided by true negative instances and the instance classifier. If the 
current instance $x_{i,j}$ comes from a positive bag, we use its embedding $q_{i,j}$ and the prototype vectors ${\mu}_r$ to generate its pseudo label $s_{i,j}\in\mathbb{R}^2$. 
At the same time, we update the prototype vector of the corresponding class using its predicted label ${\hat{y}}_{i,j}$ and embedding $q_{i,j}$. If the current instance $x_{i,j}$ 
comes from a negative bag, we directly assign it a negative label and update the negative prototype vector using its embedding $q_{i,j}$. Then, we use the generated pseudo labels 
to train the instance classifier and complete this iteration.

\textbf{Pseudo Label Generation.} If the current instance $x_{i,j}$ comes from a positive bag, we calculate the inner product between its embedding $q_{i,j}$ and the two 
prototype vectors ${\mu}_r$, and select the prototype label with the smaller feature distance as the update direction $z_{i,j}\in\mathbb{R}^2$ for the pseudo 
label of $x_{i,j}$. Then, we use a moving updating strategy to update the pseudo label of the instance, defined as follows:
\begin{equation}
   s_{i, j}=\alpha s_{i, j}+(1-\alpha) z_{i, j}, z_{i, j}=onehot\left(\operatorname{argmax} q_{i, j}{ }^{\top} {\mu}_r\right),  \label{eq6}
\end{equation}
where $\alpha$ is a coefficient for moving updating, and $onehot(\cdot)$ is a function that converts a value to a two-dimensional one-hot vector. 
The moving updating strategy can make the process of updating pseudo labels smoother and more stable.

\textbf{Prototype Updating.} If the current instance $x_{i,j}$ comes from a positive bag, we update the corresponding prototype vector ${\mu}_c$ according to its 
predicted category ${\hat{y}}_{i,j}$ and embedding $q_{i,j}$ using a moving updating strategy as follows:
\begin{equation}
   {\mu}_c=\mathit{Norm}\left(\beta {\mu}_c+(1-\beta) q_{i, j}\right), c=\operatorname{argmax} \hat{y}_{i, j},  \label{eq7}
\end{equation}
where $\beta$ is a coefficient for moving updating and $\mathit{Norm}(\cdot)$ is the normalization function.

If the current instance $x_{i,j}$ comes from a negative bag, \textit{i.e.}, $x_{i,j}$ is a true negative instance, we update the negative prototype vector ${\mu}_0$ using its embedding $q_{i,j}$ as follows:
\begin{equation}
   {\mu}_0=\mathit{Norm}\left(\beta {\mu}_0+(1-\beta) q_{i, j}\right)  \label{eq8}
\end{equation}

\textbf{Instance Classification Loss.} We use the cross-entropy loss between the predicted value $p_{i,j}\in\mathbb{R}^2$ of the instance classifier and the pseudo label $s_{i,j}$ to train the instance classifier.
\begin{equation}
   \mathcal{L}_{c l s}=\mathit{C E}\left(p_{i, j}, s_{i, j}\right),  \label{eq9}
\end{equation}
where $\mathit{CE}(\cdot)$ represents the cross-entropy loss function.

\subsection{Bag Constraint and Total Loss}
\label{sec35}

\textbf{Bag Constraint.} To further utilize the bag labels, we record the bag index of each instance and apply the following bag constraint loss:
\begin{equation}
   \mathcal{L}_{b c}=\mathit{C E}\left(\mathit{M L P}\left({Mean}\left(q_{i, j}, j=1,2, \ldots n_i\right)\right), Y_i\right),  \label{eq10}
\end{equation}
where $Mean(q_{i,j},j=1,2,\ldots n_i)$ represents the mean pooling of all instance embeddings in a bag to obtain a bag embedding.

\textbf{Total Loss.} The total loss $\mathcal{L}$ is composed of the contrastive loss $\mathcal{L}_{\mathit{IWSCL\ }}$, instance classification loss $\mathcal{L}_{cls}$, and bag constraint loss $\mathcal{L}_{bc}$, defined as follows:
\begin{equation}
   \mathcal{L}=\mathcal{L}_\mathit{I W S C L}+\lambda_1 \mathcal{L}_{c l s}+\lambda_2 \mathcal{L}_{b c},  \label{eq11}
\end{equation}
where $\lambda_1$ and $\lambda_2$ are weight coefficients used for balancing.

\section{Experimental Results}
\subsection{Datasets}
We used four datasets to comprehensively evaluate the instance and bag classification performance of INS, including a simulated dataset called CIFAR-MIL, constructed using CIFAR 10 \cite{1}, as well as three real WSI datasets: the Camelyon 16 Dataset \cite{3} for breast cancer, the TCGA-Lung Cancer Dataset, and an in-house Cervical Cancer Dataset. More importantly, our experiments not only include the tasks that doctors can directly judge from H\&E stained slides, including tumor diagnosis (on the Camelyon 16 Dataset) and tumor subtyping (on the TCGA-Lung Cancer Dataset), but also the tasks that doctors cannot directly make decisions from HE slides, including lymph node metastasis from primary lesion, patient prognosis, and prediction of immunohistochemical marker (all on the Cervical Cancer Dataset).
\subsubsection{Simulated CIFAR-MIL Dataset}
To evaluate the performance of INS under different positive ratios and compare it with the comparison methods, following WENO \cite{2}, we used 10-class natural image datasets CIFAR-10 \cite{1} to construct a simulated WSI dataset called CIFAR-MIL with different positive ratios.

The CIFAR-10 dataset consists of 60,000 32$\times$32-pixel color images divided into 10 categories (airplane, automobile, bird, cat, deer, dog, frog, horse, ship, truck), with each category containing 6,000 images. Out of these images, 50,000 are used for training and 10,000 for testing. To simulate pathological Whole Slide Images (WSIs), we combined a set of random images from each category of the CIFAR-10 dataset to construct the WSI. Specifically, we treated each image from each category of the CIFAR-10 dataset as an instance, and only all instances of the "truck" category were labeled as positive while the other instances were labeled as negative (the truck category is chosen at random). Then, we randomly selected a positive bag consisting of a set of $a$ positive instances and $100-a$ negative instances (without repetition) from all instances, with a positive ratio of $\frac{a}{100}$. Similarly, we randomly selected a negative bag consisting of 100 negative instances (without repetition). We repeated this process until all positive or negative instances in the CIFAR-10 dataset were used up. By adjusting the value of a, we constructed 5 subsets of the CIFAR-MIL dataset with positive ratios of 5\%, 10\%, 20\%, 50\%, and 70\%, respectively. 

\subsubsection{Camelyon16 Public Dataset}
The Camelyon16 dataset is a publicly available dataset of histopathology images used for detecting breast cancer metastasis in lymph nodes \cite{3}. WSIs containing metastasis are labeled positive, while those without metastasis are labeled negative. In addition to slide-level labels indicating whether a WSI is positive or negative, the dataset also provides pixel-level labels for metastasis areas. To satisfy weakly supervised scenarios, we used only slide-level labels for training and evaluated the instance classification performance of each algorithm using the pixel-level labels of cancer areas. Prior to training, we divided each WSI into non-overlapping 512$\times$512 image patches under 10$\times$ magnification. Patches with entropy less than 5 were removed as background, and a patch was labeled positive if it contained 25\% or more cancer areas, otherwise, it was labeled negative. A total of 186,604 instances were obtained for training and testing, with 243 slides used for training and 111 for testing.

\subsubsection{TCGA Lung Cancer Dataset}
The TCGA Lung Cancer dataset comprises 1054 WSIs obtained from the Cancer Genome Atlas (TCGA) Data Portal, which consists of two lung cancer subtypes, namely Lung Adenocarcinoma and Lung Squamous Cell Carcinoma. Our objective is to accurately diagnose both subtypes, with WSIs of Lung Adenocarcinoma labeled as negative and WSIs of Lung Squamous Cell Carcinoma labeled as positive. This dataset provides only slide-level labels and patch-level labels are unavailable. The dataset contains about 5.2 million patches at 20$\times$ magnification, with an average of approximately 5,000 patches per slide. These WSIs were randomly partitioned into 840 training slides and 210 test slides (4 low-quality corrupted slides are discarded). 

\subsubsection{Cervical Cancer Dataset}
The Cervical Cancer dataset is an in-house clinical pathology dataset that includes 374 H\&E-stained WSIs of primary lesions of cervical cancer from different patients, after slide selection. 
We conducted all experiments at 5$\times$ magnification, and we divided each WSI into non-overlapping patches of size 224$\times$224 to form a bag. Background patches with entropy values less than 5 were discarded from the original WSI. 
For \textbf{prediction of lymph node metastasis in primary tumors}, we labeled the corresponding slides of patients who developed pelvic lymph node metastasis as positive (209 cases) and those who did not develop pelvic lymph node metastasis as negative (165 cases). We randomly divided the WSIs into a training set (300 cases) and a test set (74 cases).
For \textbf{prediction of patient survival prognosis}, following Skrede \textit{et al.} \cite{6}, we grouped all patients based on detailed follow-up records using the median as a cutoff, where those who did not experience cancer-related death within three years were labeled as negative (good prognosis) and those who did were labeled as positive (poor prognosis). Then, we randomly divided the WSIs into a training set (294 cases) and a test set (80 cases) according to the labels.
For \textbf{prediction of the immunohistochemical marker KI-67}, following Liu \textit{et al.} \cite{7} and Feng \textit{et al.} \cite{8}, we grouped all patients based on detailed KI-67 immunohistochemistry reports using the median as a cutoff, where KI-67 levels below 75 were labeled as negative and those above 75 were labeled as positive. Then, we randomly divided the WSIs into a training set (294 cases) and a test set (80 cases) according to the labels.

\subsection{Evaluation Metrics and Comparison Methods}
For both instance and bag classification, we used Area Under Curve (AUC) and Accuracy as evaluation metrics. Bag classification performance is evaluated on all datasets but instance-level classification performance is only evaluated on the CIFAR-MIL Dataset and the Camelyon 16 Dataset, since only these two datasets have instance labels.
We compared our INS to 11 competitors, including three instance-based methods: MILRNN\cite{1}, Chi-MIL\cite{2}, and DGMIL\cite{3}, and eight bag-based methods:  ABMIL\cite{8}, Loss-ABMIL\cite{12}, CLAM \cite{37}, DSMIL\cite{13}, TransMIL\cite{15}, DTFD-MIL\cite{14}, TPMIL \cite{91} and WENO\cite{18}. In accordance with DSMIL \cite{4}, we employed SimCLR \cite{5} as the self-supervised approach to pre-extract patch features for all techniques. Regarding all comparative approaches, we replicated them using the published codes and conducted a grid search on the crucial hyperparameters across all methods. We cited the reported results from their papers under the same experimental settings.

\subsection{Implementation Details}
In line with DSMIL \cite{4}, we conducted pre-processing on WSI datasets such as patch cropping and background removal.
For all datasets, the encoders are implemented using the ResNet18. The Instance classifier is implemented using MLP. The Projector is a 2-layer MLP and the Prototype vectors are 128 dimensions. No pre-training of the network parameters is performed. The SGD optimizer is used to optimize the network parameters with a learning rate of 0.01, momentum of 0.9 and the batch size is 64. The length of Embedding Queue is 8192. In order to smooth the training process, we empirically set up the warm-up epochs. After warm-up, we updated pseudo labels and gave true negative labels. The hyperparameter thresholds vary for each dataset, and we used grid search on the validation set to determine the optimal values. 
For more details, please refer to our codes, which will be available soon.

\subsection{Results on the Synthetic Dataset CIFAR-MIL}
We constructed the synthetic WSI datasets CIFAR-MIL with varying positive instance ratios to investigate the instance and bag classification performance of each method at different positive instance ratios. The results are shown in Table \ref{table1} and Table \ref{table2}. It can be seen that INS achieved the best performance in both instance and bag classification 
tasks at all positive instance ratios. Most methods cannot work well in low positive instance ratios of 5\% and 10\%, for which the AUC of INS exceeds 0.94. At positive instance ratios 
above 20\%, the performance of INS is comparable to that of fully supervised methods.

Another interesting phenomenon is that most methods have higher bag classification performance when the positive ratio is higher than 20\%, but the instance classification 
performance is still fairly poor. This suggests that accurately identifying all positive instances is not necessary to complete correct bag classification. As the 
positive ratio increases, there are many positive instances in positive bags, which makes the bag classification task easier. The network often only needs to 
identify the simplest positive instances to complete bag classification, losing the motivation to accurately classify all positive instances. In contrast, INS 
maintains high instance and bag classification performance at all positive instance ratios.

% Table 1
\begin{table}[t!]
% \vspace{-0.7em}
\centering
\caption{Instance classification results on the CIFAR-MIL Dataset.}
 \scalebox{0.9}{
\begin{tabular}{
>{\columncolor[HTML]{FFFFFF}}c 
>{\columncolor[HTML]{FFFFFF}}c 
>{\columncolor[HTML]{FFFFFF}}c 
>{\columncolor[HTML]{FFFFFF}}c 
>{\columncolor[HTML]{FFFFFF}}c 
>{\columncolor[HTML]{FFFFFF}}c }
\toprule[1pt]
Positive instance ratio                               & 5\%                           & 10\%                          & 20\%                          & 50\%                          & 70\%                          \\
 \midrule[1pt]
{\color[HTML]{808080} Fully supervised} & {\color[HTML]{808080} 0.9621} & {\color[HTML]{808080} 0.9723} & {\color[HTML]{808080} 0.9740} & {\color[HTML]{808080} 0.9699} & {\color[HTML]{808080} 0.9715} \\ \hline
ABMIL (18'ICML)                         & 0.8485                        & 0.8505                        & 0.8909                        & 0.8224                        & 0.7935                        \\
Loss-ABMIL (20'AAAI)                    & 0.6915                        & 0.7372                        & 0.7430                        & 0.7475                        & 0.6881                        \\
Chi-MIL (20'MICCAI)                     & 0.5872                        & 0.6801                        & 0.7039                        & 0.6851                        & 0.7091                        \\
DSMIL (21'CVPR)                          & 0.5515                        & 0.4918                        & 0.8258                        & 0.6152                        & 0.7525                        \\
DGMIL (22'MICCAI)                       & 0.7016                        & 0.7818                        & 0.8815                        & 0.8217                        & 0.8308                        \\
WENO (22'NeurIPS)                       & 0.9408                        & 0.9179                        & 0.9657                        & 0.9393                        & 0.9525                        \\ 
\hline
\textbf{INS (ours)}                     & {\color[HTML]{FE0000} \textbf{0.9418}}               & {\color[HTML]{FE0000} \textbf{0.9466}}               & {\color[HTML]{FE0000} \textbf{0.9668}}               & {\color[HTML]{FE0000} \textbf{0.9702}}               & {\color[HTML]{FE0000} \textbf{0.9720} }              \\
\bottomrule[1pt]
\end{tabular}}
  \label{table1}
\end{table}
% \vspace{-0.4em}
% Table 2
\begin{table}[t!]
% \vspace{-1.5em}
\centering
\caption{Bag classification results on the CIFAR-MIL Dataset.}
 \scalebox{0.9}{
\begin{tabular}{
>{\columncolor[HTML]{FFFFFF}}c 
>{\columncolor[HTML]{FFFFFF}}c 
>{\columncolor[HTML]{FFFFFF}}c 
>{\columncolor[HTML]{FFFFFF}}c 
>{\columncolor[HTML]{FFFFFF}}c 
>{\columncolor[HTML]{FFFFFF}}c }
\toprule[1pt]
Positive instance ratio                               & 5\%                           & 10\%                          & 20\%                          & 50\%                          & 70\%                          \\
 \midrule[1pt]
{\color[HTML]{808080} Fully supervised} & {\color[HTML]{808080} 0.9621} & {\color[HTML]{808080} 0.9723} & {\color[HTML]{808080} 0.9740} & {\color[HTML]{808080} 0.9699} & {\color[HTML]{808080} 0.9715} \\ 
\hline
ABMIL (18'ICML)                         & 0.6783                                 & 0.9344                                 & 0.9678                                 & 1.0000                                 & \textbf{1.0000}                        \\
Loss-ABMIL (20'AAAI)                    & 0.4913                                 & 0.9108                                 & 0.9352                                 & 0.9475                                 & \textbf{1.0000}                        \\
Chi-MIL (20'MICCAI)                     & 0.5630                                 & 0.4519                                 & 0.8633                                 & 0.9105                                 & \textbf{1.0000}                        \\
DSMIL(21'CVPR)                          & 0.5174                                 & 0.5265                                 & 0.9468                                 & 0.9850                                 & \textbf{1.0000}                        \\
DGMIL (22'MICCAI)                       & 0.7159                                 & 0.9414                                 & 0.9681                                 & 0.9838                                 & \textbf{1.0000}                        \\
WENO (22'NeurIPS)                       & 0.9367                                 & 0.9900                                 & \textbf{1.0000}                        & \textbf{1.0000}                        & \textbf{1.0000}                        \\ \hline
\textbf{INS (ours)}                     & {\color[HTML]{FE0000} \textbf{0.9408}} & {\color[HTML]{FE0000} \textbf{0.9903}} & {\color[HTML]{FE0000} \textbf{1.0000}} & {\color[HTML]{FE0000} \textbf{1.0000}} & {\color[HTML]{FE0000} \textbf{1.0000}} \\ 
\bottomrule[1pt]
\end{tabular}}
 \label{table2}
\end{table}
% \vspace{-1.1em}

\begin{table}[t!]
   \centering
   \caption{Results on the Camelyon 16 Dataset.}
   \setlength{\tabcolsep}{2.6mm}{
   \scalebox{0.8}{
   \begin{tabular}{
   >{\columncolor[HTML]{FFFFFF}}c 
   >{\columncolor[HTML]{FFFFFF}}c 
   >{\columncolor[HTML]{FFFFFF}}c 
   >{\columncolor[HTML]{FFFFFF}}c 
   >{\columncolor[HTML]{FFFFFF}}c l}
   \toprule[1pt]
   \cellcolor[HTML]{FFFFFF}                          & \multicolumn{2}{c}{\cellcolor[HTML]{FFFFFF}Instance-level}                      & \multicolumn{2}{c}{\cellcolor[HTML]{FFFFFF}Bag-level}                           &  \\
   \multirow{-2}{*}{\cellcolor[HTML]{FFFFFF}Methods} & AUC                                    & Accuracy                               & AUC                                    & Accuracy                               &  \\ 
   \midrule[1pt]
   ABMIL (18'ICML)                                   & 0.8480                                 & 0.8033                                 & 0.8379                                 & 0.8198                                 &  \\
   MILRNN (19'Nat. Med)                              & 0.8568                                 & 0.8174                                 & 0.8262                                 & 0.8198                                 &  \\
   Loss-ABMIL (20'AAAI)                              & 0.8995                                 & 0.8512                                 & 0.8299                                 & 0.8018                                 &  \\
   Chi-MIL (20'MICCAI)                               & 0.7880                                 & 0.7453                                 & 0.8256                                 & 0.8018                                 &  \\
   DSMIL (21'CVPR)                                   & 0.8858                                 & 0.8566                                 & 0.8401                                 & 0.8108                                 &  \\
   TransMIL (21'NeurIPS)                             & -                                      & -                                      & 0.8566                                 & 0.8288                                 &  \\
   CLAM (21' Nat. Biomed. Eng.)                      & 0.8913                                 & 0.8655                                 & 0.8594                                 & 0.8378                                 &  \\
   DTFD-MIL (22'CVPR)                                & 0.8928                                 & 0.8701                                 & 0.8638                                 & 0.8468                                 &  \\
   DGMIL (22'MICCAI)                                 & 0.9012                                 & 0.8859                                 & 0.8368                                 & 0.8018                                 &  \\
   WENO (22'NeurIPS)                                 & 0.9377                                 & 0.9057                                 & 0.8495                                 & 0.8108                                 &  \\
   TPMIL (23'MIDL)                                   & 0.9234                                 & 0.8867                                 & 0.8421                                 & 0.8018                                 &  \\ 
   \hline
   \textbf{INS (ours)}                               & {\color[HTML]{FF0000} \textbf{0.9583}} & {\color[HTML]{FF0000} \textbf{0.9249}} & {\color[HTML]{FF0000} \textbf{0.9016}} & {\color[HTML]{FF0000} \textbf{0.8739}} &  \\ 
   \bottomrule[1pt]   
   \end{tabular}}}
   \label{table3}
\end{table}
   % \vspace{-1.5em}

% Table 4
\begin{table}[t!]
   \centering
%    \vspace{-0.5em}
   \caption{Results on the TCGA-Lung Cancer Dataset.}
   \setlength{\tabcolsep}{8.5mm}{
   \scalebox{0.8}{
   \begin{tabular}{
   >{\columncolor[HTML]{FFFFFF}}c 
   >{\columncolor[HTML]{FFFFFF}}c 
   >{\columncolor[HTML]{FFFFFF}}c}
   \toprule[1pt]
   \cellcolor[HTML]{FFFFFF}                          & \multicolumn{2}{c}{\cellcolor[HTML]{FFFFFF}Bag-level}                           \\ 
   \multirow{-2}{*}{\cellcolor[HTML]{FFFFFF}Methods} & AUC                                    & Accuracy                               \\ 
   \midrule[1pt]
   ABMIL (18'ICML)                                   & 0.9488                                 & 0.9000                                 \\
   MILRNN (19'Nat. Med)                              & 0.9107                                 & 0.8619                                 \\
   Loss-ABMIL (20'AAAI)                              & 0.9517                                 & 0.9143                                 \\
   Chi-MIL (20'MICCAI)                               & 0.9523                                 & 0.9190                                 \\
   DSMIL(21'CVPR)                                    & 0.9633                                 & 0.9190                                 \\
   TransMIL(21'NeurIPS)                              & 0.9830                                 & 0.9381                                 \\
   CLAM (21'Nat. Biomed. Eng.)                                 & 0.9788                                 & 0.9286                                 \\
   DTFD-MIL(22'CVPR)                                 & 0.9808                                 & 0.9524                                 \\
   DGMIL (22'MICCAI)                                 & 0.9702                                 & 0.9190                                 \\
   WENO (22'NeurIPS)                                 & 0.9727                                 & 0.9238                                 \\ 
   TPMIL (23'MIDL)                                 & 0.9799                                 & 0.9429                                 \\\hline
   \textbf{INS (ours)}                               & {\color[HTML]{FF0000} \textbf{0.9837}} & {\color[HTML]{FF0000} \textbf{0.9571}} \\
    \bottomrule[1pt]
   \end{tabular}}}
   \label{table4}
\end{table}

% Table 5
\begin{table}[t!]
   \centering
%    \vspace{-0.5em}
   \caption{Results on the Cervical Cancer Dataset.}
   \setlength{\tabcolsep}{2.4mm}{
   \scalebox{0.65}{
   \begin{tabular}{
   >{\columncolor[HTML]{FFFFFF}}c
   >{\columncolor[HTML]{FFFFFF}}c 
   >{\columncolor[HTML]{FFFFFF}}c 
   >{\columncolor[HTML]{FFFFFF}}c 
   >{\columncolor[HTML]{FFFFFF}}c 
   >{\columncolor[HTML]{FFFFFF}}c 
   >{\columncolor[HTML]{FFFFFF}}c }
   \toprule[1pt]
   Clinical Tasks (Bag level) & \multicolumn{2}{c}{\cellcolor[HTML]{FFFFFF}Lymph node metastasis}              & \multicolumn{2}{c}{\cellcolor[HTML]{FFFFFF}Survival Prognosis}                                & \multicolumn{2}{c}{\cellcolor[HTML]{FFFFFF}KI-67 Prediction}                    \\ \hline
   Methods                    & AUC                                    & Accuracy                               & AUC                                    & \multicolumn{1}{c}{\cellcolor[HTML]{FFFFFF}Accuracy} & AUC                                    & Accuracy                               \\ 
    \midrule[1pt]
   ABMIL (18'ICML)            & 0.7716                                 & 0.7297                                 & 0.7439                                 & \multicolumn{1}{c}{\cellcolor[HTML]{FFFFFF}0.7125}   & 0.6963                                 & 0.6500                                 \\
   Loss-ABMIL (20'AAAI)       & 0.7835                                 & 0.7568                                 & 0.7505                                 & \multicolumn{1}{c}{\cellcolor[HTML]{FFFFFF}0.7250}   & 0.6987                                 & 0.6625                                 \\
   Chi-MIL (20'MICCAI)        & 0.7783                                 & 0.7432                                 & 0.7349                                 & \multicolumn{1}{c}{\cellcolor[HTML]{FFFFFF}0.7000}   & 0.6933                                 & 0.6625                                 \\
   DSMIL (21'CVPR)            & 0.8022                                 & 0.7838                                 & 0.7622                                 & \multicolumn{1}{c}{\cellcolor[HTML]{FFFFFF}0.7375}   & 0.7124                                 & 0.6875                                 \\
   TransMIL (21'NeurIPS)      & 0.8126                                 & 0.7838                                 & 0.7716                                 & \multicolumn{1}{c}{\cellcolor[HTML]{FFFFFF}0.7375}   & 0.7252                                 & 0.6875                                 \\
   CLAM (21'Nat. Biomed. Eng.)        & 0.8138                                 & 0.7838                                 & 0.7755                                 & \multicolumn{1}{c}{\cellcolor[HTML]{FFFFFF}0.7375}   & 0.7297                                 & 0.7000                                 \\
   DTFD-MIL (22'CVPR)         & 0.8108                                 & 0.7703                                 & 0.7650                                 & \multicolumn{1}{c}{\cellcolor[HTML]{FFFFFF}0.7375}   & 0.7267                                 & 0.6875                                 \\
   DGMIL (22'MICCAI)          & 0.8056                                 & 0.7568                                 & 0.7515                                 & \multicolumn{1}{c}{\cellcolor[HTML]{FFFFFF}0.7250}   & 0.7115                                 & 0.6875                                 \\
   WENO (22'NeurIPS)          & 0.8222                                 & 0.7703                                 & 0.7775                                 & \multicolumn{1}{c}{\cellcolor[HTML]{FFFFFF}0.7500}   & 0.7323                                 & 0.7000                                 \\ 
   TPMIL (23'MIDL)            & 0.8213                                 & 0.7703                                 & 0.7728                                 & \multicolumn{1}{c}{\cellcolor[HTML]{FFFFFF}0.7500}   & 0.7318                                 & 0.7000                                 \\ \hline
   \textbf{INS (ours)}        & {\color[HTML]{FF0000} \textbf{0.8677}} & {\color[HTML]{FF0000} \textbf{0.8243}} & {\color[HTML]{FF0000} \textbf{0.7812}} & {\color[HTML]{FF0000} \textbf{0.7625}}                & {\color[HTML]{FF0000} \textbf{0.7534}} & {\color[HTML]{FF0000} \textbf{0.7125}} \\ 
    \bottomrule[1pt]
   \end{tabular}}}
    % \vspace{-0.5em}
   \label{table5}
\end{table}

\subsection{Results on Real-World Datasets}

\subsubsection{Camelyon 16 Dataset}
\begin{figure}[t!]
   \begin{center}
   \includegraphics[width=0.9\linewidth]{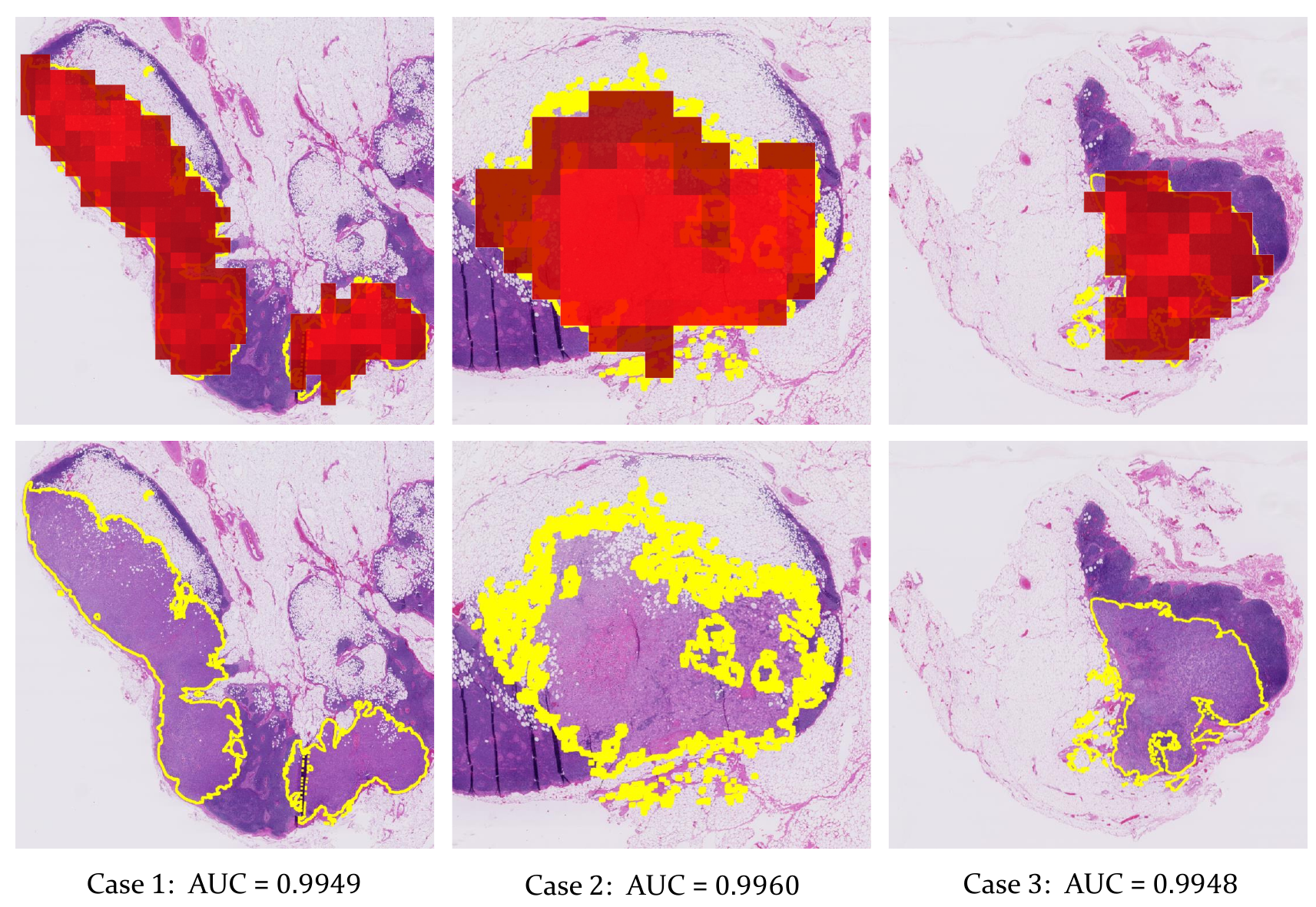}
   \end{center}
       %  \vspace{-0.9em}
      \caption{Typical visualization results on the Camelyon 16 Dataset, where the yellow line represents the true tumor boundary annotated by doctors, and the pick boxes represent 
      the positive instances predicted by INS as heatmaps.}
   \label{figure4}
\end{figure}

Table \ref{table3} shows the classification performance of INS and other methods on the Camelyon 16 Dataset. The low positive ratio of Camelyon 16 Dataset (about 10\%-20\%) makes 
the classification task quite difficult. INS outperforms all the compared methods with a large margin, exceeding the second-best method by 2.1\% and 3.8\% in AUC for instance and bag 
classification, respectively.  Figure \ref{figure4} shows typical visualization results on this dataset. It can be seen that INS accurately localizes almost all positive instances in the positive bags.

\subsubsection{TCGA-Lung Cancer Dataset}
Table \ref{table4} shows the results on the TCGA-Lung Cancer Dataset. Unlike the Camelyon 16 Dataset, this dataset has a high positive instance ratio 
(more than 80\%), so the performance of all methods is fairly good, while INS still achieves the best performance.

\subsubsection{Cervical Cancer Dataset}
Table \ref{table5} shows the experimental results of three tasks on the in-house Cervical Cancer Dataset, including lymph node metastasis prediction from primary lesion, 
patient prognosis, and immunohistochemical marker KI-67 prediction from H\&E stained WSIs. Unlike the previous two real-world datasets, these three tasks are very tough 
even for human doctors. It can be seen that INS significantly outperforms other methods in all three tasks, demonstrating the strong performance of INS.

\begin{figure}[t!]
   \begin{center}
   \includegraphics[width=0.9\linewidth]{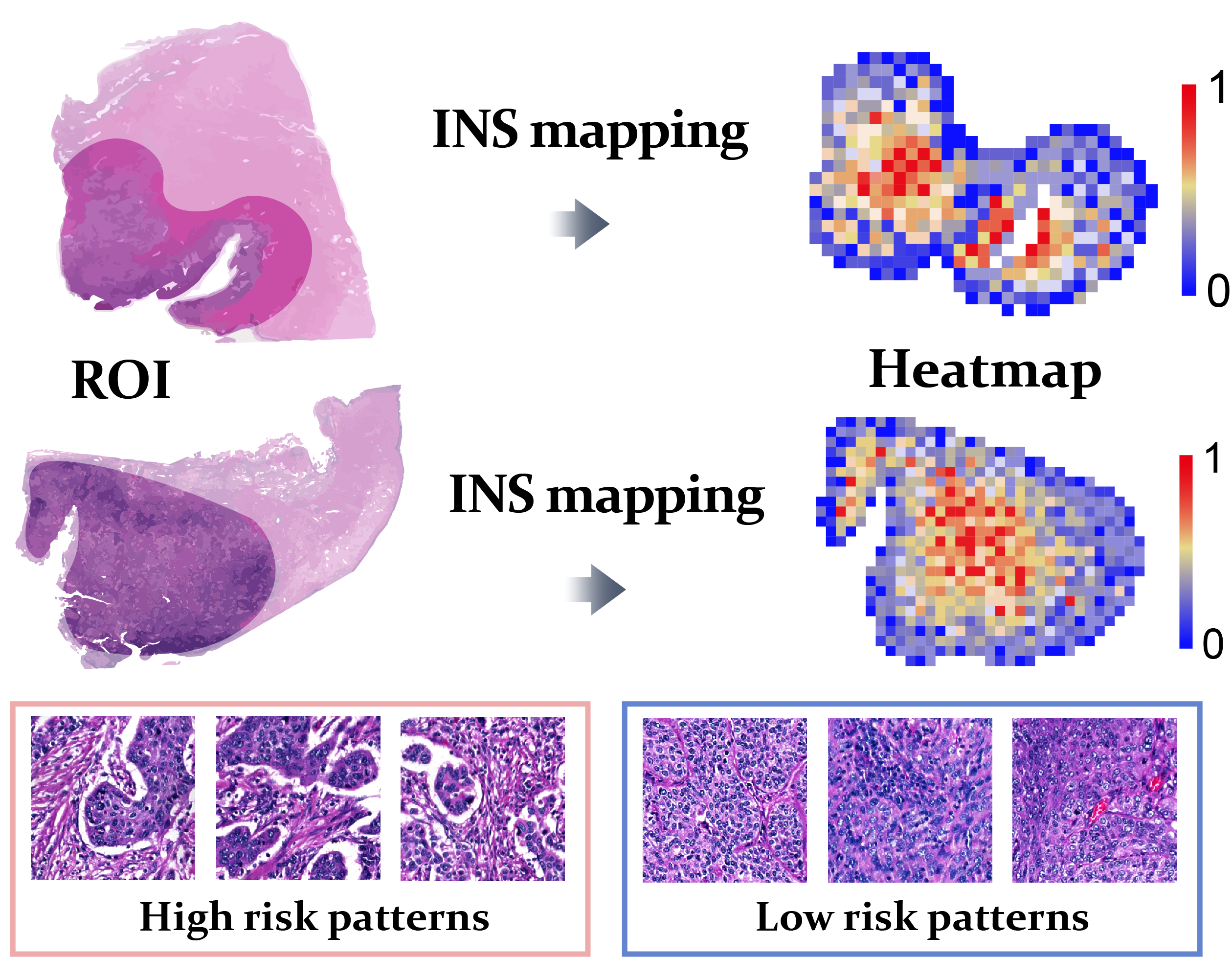}
   \end{center}
    %  \vspace{-1em}
      \caption{Heatmap visualization of INS on the lymph node metastasis task, where we identify the "Micropapillae" pathological pattern that indicates positive metastasis from HE slides.}
    %  \vspace{-1.75em}
   \label{figure5}
\end{figure}

\subsection{Interpretability Study of the Lymph Node Metastasis Task}
As can be seen from Table \ref{table5}, INS can predict the preoperative lymph node status of cervical cancer with high performance using HE-stained slides of the primary tumor. However, this does not offer pathologists additional interpretable pathological patterns. While cancer diagnosis is possible through visual inspection of HE-stained pathological slides, the same cannot be said for accurately determining lymph node metastasis status from the primary lesion. Given INS’s powerful instance classification ability, we hope to find explicit pathological patterns that can suggest the lymph node metastasis status and perform interpretable analysis as well as new knowledge discovery.
Specifically, we used INS to predict the probability of each instance being positive within the positive bags, and visualized the top 0.1\% instances with the highest and lowest probabilities separately. These instances with the highest predicted positive probabilities are likely to contain pathological patterns that identify high-risk lymph node metastasis, as shown in Figure \ref{figure5}.

As can be seen, in images suggesting lymph node metastasis, structures resembling "micropapillae" are more prevalent, indicating a high-risk pathological pattern. Micropapillary structures are characterized by small clusters of infiltrating cancer cells forming hollow or mulberry-like nests without a central fibrovascular axis, surrounded by blank lacunae or lacunae between interstitial components. Conversely, negative lymph node images more commonly exhibit a "sheet-like" pattern, in which cells form tightly connected nests of varying sizes within the tumor interstitium, with fissures or lacunae rarely observed.
This conclusion highlights the interpretability of using INS to assess lymph node metastasis and its significant guiding implications for future clinical research.

% Moreover, given INS's powerful instance classification ability, we can use it for interpretability research and new knowledge discovery in clinical difficult tasks. 
% Figure \ref{figure5} shows the heat map of INS results and several typical high-risk patches and low-risk patches in the lymph node metastasis task, 
% where we identify the "Micropapillae" pathological pattern that indicates high risk of lymph node metastasis. More details can be found in the Supplementary Material.
% \vspace{-0.5em}
% figure4
%------------------------------------------------------------------------
\section{Ablation Study and Further Analysis}
\subsection{Ablation Study on Key Components}
We conducted comprehensive ablation experiments on the components of INS using the Camelyon 16 Dataset, and the results are shown in Table \ref{table6}. 
Here, w/o $\mathcal{L}_{\mathit{IWSCL\ }}$ means that weakly supervised contrastive learning is not used, w/o $\mathcal{L}_{bc}$ means that bag-level 
constraint is not used, and w/o MU means that $\alpha$=0 in formula \ref{eq4}, indicating that the pseudo label updating strategy with moving updating is not used. 
It can be seen that IWSCL is the crucial component of INS, and without contrastive learning, the performance of INS declines significantly. 
Both the bag constraint and the pseudo label updating strategy with moving updating can effectively improve the performance of INS. 
It is worth noting that INS still outperforms all comparing methods even without either one of these two components.

% Table 6
\begin{table}[t!]
\centering
%  \vspace{-0.5em}
\caption{Results of ablation study on the Camelyon 16 Dataset.}
 \scalebox{0.9}{
\begin{tabular}{
>{\columncolor[HTML]{FFFFFF}}c 
>{\columncolor[HTML]{FFFFFF}}c 
>{\columncolor[HTML]{FFFFFF}}c 
>{\columncolor[HTML]{FFFFFF}}c 
>{\columncolor[HTML]{FFFFFF}}c 
>{\columncolor[HTML]{FFFFFF}}c }
 \toprule[1pt]
Conditions    & $\mathcal{L} _\mathit{IWSCL}$  & $\mathcal{L}_{bc}$ & pseudo label & Instance AUC                           & Bag AUC                                \\  \midrule[1pt]
\textbf{Ours} & \checkmark        & \checkmark     & \checkmark            & {\color[HTML]{FF0000} \textbf{0.9583}} & {\color[HTML]{FF0000} \textbf{0.9016}} \\
w/o  $\mathcal{L} _\mathit{IWSCL}$  &          & \checkmark     & \checkmark            & 0.9005                                 & 0.8301                                 \\
w/o $\mathcal{L}_{bc}$     & \checkmark        &       & \checkmark            & 0.9411                                 & 0.8734                                 \\
w/o MU        & \checkmark        & \checkmark     & \checkmark            & 0.9423                                 & 0.8815                                 \\ 
 \bottomrule[1pt]
\end{tabular}}
%  \vspace{-1em}
\label{table6}
\end{table}

\subsection{Further Analysis}
\label{FA}
We conducted more detailed evaluation and interpretive experiments of INS.

\begin{figure}[t!]
   \begin{center}
   \includegraphics[width=1 \linewidth]{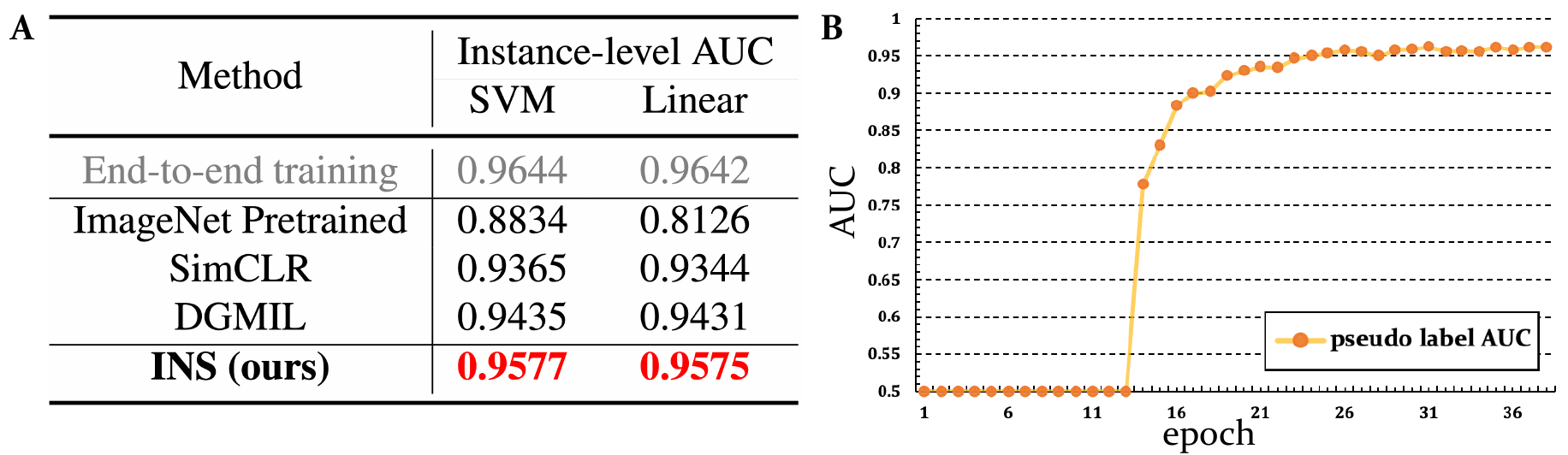}
   \end{center}
    %  \vspace{-1.2em}
    \caption{A. SVM and linear evaluation of pre-extracted features on the Camelyon16 Dataset. B. Visualization of the quality assessment of generated pseudo labels.}
    %  \vspace{-1em}
   \label{figure6}
\end{figure}

\textbf{Effective Representation Learning of INS.} Currently, many studies \cite{13,14,15,18,40,41} use pre-trained feature extractors to extract instance features for subsequent training, 
among which ImageNet pretrained method \cite{15} and contrastive self-supervised methods \cite{13,14,18,41,43} are most commonly used. We compared these feature extraction methods with 
INS on the Camelyon16 Dataset. Specifically, we first used feature extractors pre-trained by ImageNet \cite{50}, SimCLR \cite{31}, DGMIL \cite{3}, and our INS, respectively, to extract all instance features. 
All methods used ResNet-18 as the network structure. Then we used the true labels of each instance only based on these features to train a simple SVM classifier and a linear classifier and 
tested the two classifiers on the test set to evaluate these features. Please note that the pre-training process of 
feature extractors is unsupervised (not using labels) or weakly supervised (using only bag labels), while the process of training SVM and Linear with the pre-extracted features is 
fully supervised using instance labels, enabling an effective evaluation of the quality of the pre-extracted features. We also trained the network in an end-to-end way as an upper 
bound. The results are shown in Figure \ref{figure6} A. The features extracted from INS achieve the best results, indicating effectiveness of our method.

\begin{figure}[t!]
   \begin{center}
   \includegraphics[width=1\linewidth]{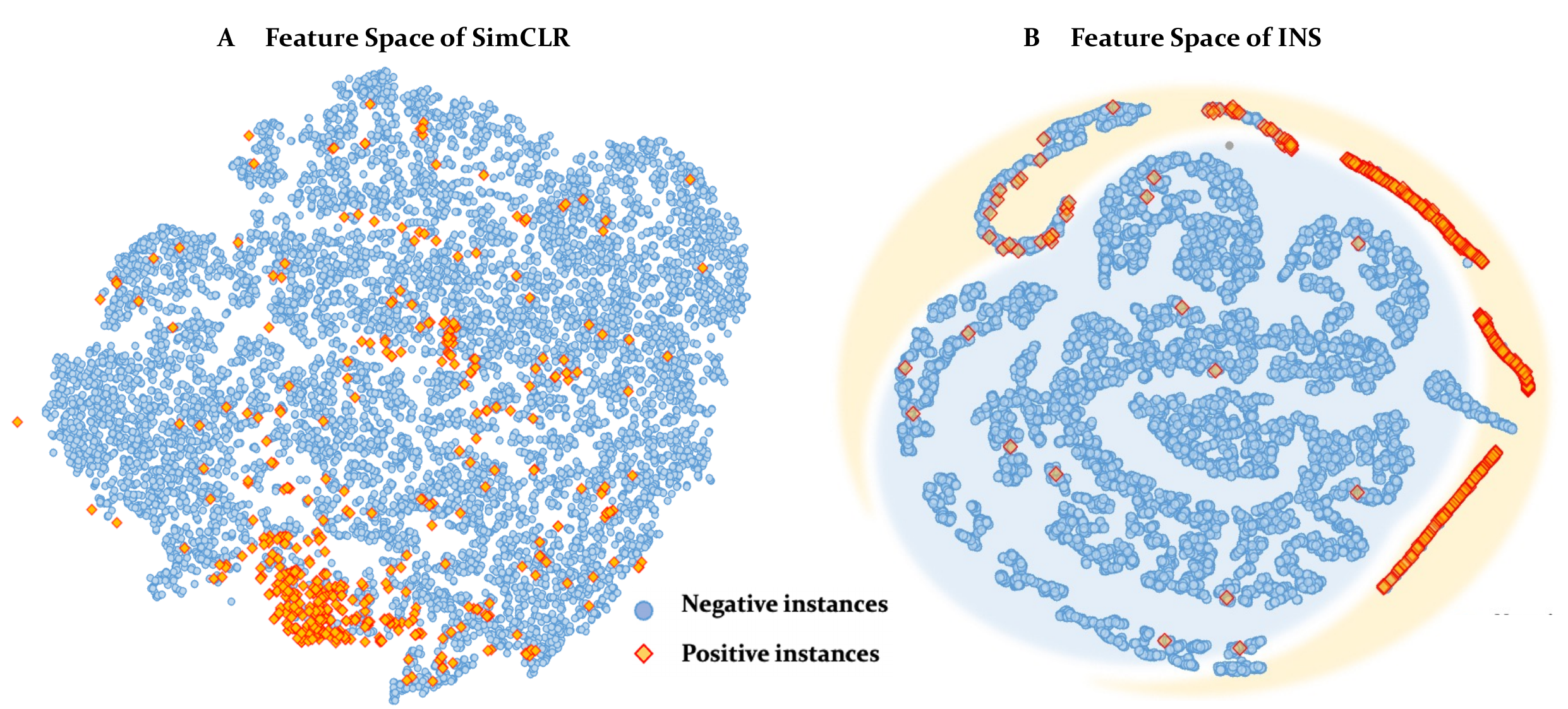}
   \end{center}
    %  \vspace{-1em}
     \caption{Visualization of feature distribution on a typical slide on the Camelyon16 Dataset using t-SNE. A. Instance features extracted by SimCLR. B. Instance features extracted by our INS.}
    %  \vspace{-1.5em}
   \label{figure7}
\end{figure}

In addition, we use SimCLR and our INS to extract features of all instances from a typical slide in the Camelyon16 Dataset and visualize the feature distribution using 
t-SNE, as shown in Figure \ref{figure7}. It can be visually observed that the positive and negative instances are separated in the features extracted by INS. Surprisingly, 
we can even manually draw a clear boundary between them on the two-dimensional plane, indicating the powerful feature representation ability of INS. In contrast, 
although the features of SimCLR are relatively scattered in the feature space, there is significant overlapping between the positive and negative instances, which cannot be easily separated.

\textbf{Evaluation of Pseudo Labels.} To intuitively demonstrate the quality of the instance pseudo labels, we plotted the AUC curve of the pseudo labels for all instances 
in the training set of the CIFAR-MIL Dataset (with a 0.2 positive instance ratio) against the training epochs, as shown in Figure \ref{figure6} B. It can be clearly seen that the quality 
of the pseudo labels continues to improve, providing more and more effective guidance to the instance classifier.

%-------------------------------------------------------------------------

\section{Discussion}
\subsection{Purpose and Clinical Relevance of Instance vs. Bag Classification in IWSCL.}

\textbf{Purpose and Meaning of IWSCL}: Instance-level Weakly Supervised Contrastive Learning (IWSCL) is designed to enhance the performance of MIL in the context of WSI analysis by improving the representation of instances (small patches derived from WSIs) in feature space. Traditional MIL approaches suffer from the challenge of accurately labeling each instance and identifying difficult positive instances. IWSCL addresses this by learning discriminative feature representations that better separate positive and negative instances, thereby facilitating more accurate pseudo-labeling and instance classification.

\textbf{Definition of Instances and Their Importance}: In our study, instances refer to the small patches into which a WSI is divided for analysis. Each instance represents a fraction of the larger slide, and its correct classification is crucial for understanding the slide's overall pathology. The distinction between positive (indicative of disease) and negative (healthy) instances is foundational to accurate diagnosis at both the instance and slide levels.

\textbf{Clinical Relevance of Bag vs. Instance Classification}:
Bag-Level Classification refers to the classification of the entire WSI (the "bag" in MIL terminology) as either positive or negative based on the aggregated information from its instances. Clinically, this is relevant for determining the presence or absence of disease in the slide as a whole, which is a crucial first step in diagnostic workflows.
Instance-Level Classification involves classifying each patch (instance) within a WSI. Clinically, this is crucial for identifying specific areas of interest, such as tumor regions, and understanding the heterogeneity within a slide. It is particularly relevant for detailed analyses, such as grading the severity of disease or guiding targeted biopsies.
While both levels of classification are clinically relevant, their importance varies with the diagnostic task. Bag-level classification is essential for initial disease presence screening, whereas instance-level classification plays a key role in detailed pathological assessments and understanding disease mechanisms.

To conclude, our work on IWSCL and the proposed MIL framework aims to improve both bag and instance-level classifications by addressing the limitations of existing methods. By enhancing the accuracy of instance classification, we indirectly improve bag-level classification outcomes, thus offering significant benefits for pathological diagnosis and analysis. We believe that our approach, by improving the granularity and accuracy of classifications, holds substantial promise for clinical applications, where precise localization and characterization of pathological features are critical.

\subsection{Reliability of Pseudo-labels and the Instance Classifier}
The accuracy of pseudo-labels is indeed critical in our approach, given their direct influence on the training of the instance classifier. We have devised a comprehensive strategy to ensure the reliability and accuracy of these pseudo-labels.

\textbf{Prototype-based Pseudo Label Generation (PPLG)}: Our PPLG strategy is at the heart of ensuring the precision of pseudo-labels. By maintaining two dynamic prototypes representing negative and positive instances within the feature space, we can generate high-quality pseudo-labels for each instance based on their proximity to these prototypes. This method significantly mitigates the pseudo-label noise by leveraging the global context of instance distributions rather than individual, isolated predictions.

\textbf{Contrastive Learning for Enhanced Feature Representation}: Prior to pseudo-label generation, our framework employs contrastive learning to refine the feature representations of instances. This step is crucial for ensuring that positive and negative instances are well-separated in the feature space, thereby enhancing the subsequent pseudo-label accuracy. Improved feature representation directly contributes to the reliability of PPLG and, by extension, the precision of the instance classifier.

\textbf{Joint Training Approach}: The training process of the instance classifier is intricately linked with both IWSCL for feature representation enhancement and PPLG for pseudo-label generation. This interconnected training regime ensures a synergistic improvement across all components. As the instance classifier becomes more accurate, it further refines the quality of feature representations and pseudo-labels, creating a positive feedback loop that elevates the overall system performance.

\textbf{Utilization of True Negative Instances}: An innovative aspect of our methodology is the strategic use of true negative instances from negative bags. These instances provide a stable reference point for the model, helping to anchor the pseudo-label generation process and ensure that our classifier is calibrated against known standards, thereby improving its predictive accuracy.

\textbf{Experimental results also directly validate the quality of the instance pseudo labels.} As shown in Section \ref{FA} "Evaluation of Pseudo Labels", we further plotted the AUC curve of the pseudo labels for all instances in the training set of the CIFAR-MIL Dataset (with a 0.2 positive instance ratio) against the training epochs, as shown in Figure \ref{figure6} B. It can be clearly seen that the quality of the pseudo labels continues to improve, providing more and more effective guidance to the instance classifier.

To conclude, our comprehensive approach, characterized by PPLG, contrastive learning for feature refinement, a synergistic training strategy, and the strategic use of true negatives, is meticulously designed to ensure the accuracy of pseudo-labels and the predictive reliability of the instance classifier. These mechanisms collectively address the challenges associated with pseudo-label noise and classifier training, thereby significantly enhancing the performance of our proposed framework in WSI classification tasks. Experimental results also directly validate the quality of the instance pseudo labels.

\subsection{Novelty of the Proposed IWSCL}
\textbf{Innovation in WSI Classification:}
Our innovation primarily arises from integrating established components of contrastive learning tailored specifically for weakly supervised instance-level WSI classification. While mechanisms like loss terms and momentum updates are well-known in the field of contrastive learning, their application within an instance-level weakly supervised context for WSI analysis is novel. The unique challenges presented by WSIs, such as their large size, weak annotation, detailed content, and significant variance in pathological features, necessitate specialized solutions. We address these by adapting and optimizing contrastive learning concepts to the specific requirements of WSI classification, thus advancing their application in the field of WSI analysis.

\textbf{Comparison with Existing Methods:}
As discussed in Section \ref{CL}, current methodologies typically commence with self-supervised learning on WSI patches to pretrain a feature extractor, followed by training using these features. Most utilize unsupervised contrastive learning to extract instance features but fail to specifically separate positive from negative instances. In contrast, our approach introduces IWSCL under the MIL framework for the first time, effectively distinguishing between negative and positive instance features, as illustrated in Figure \ref{figure2}.

\textbf{Role in Joint Training Strategy:}
The IWSCL framework is a crucial component of a comprehensive training strategy that also includes PPLG and an instance classifier. This tripartite approach synthesizes the strengths of each component, enhancing the model's robustness and effectiveness for both instance and bag-level classification tasks. The interaction between IWSCL, PPLG, and the instance classifier, underpinned by weak supervision and enhanced through contrastive learning, significantly improves the training process. It effectively utilizes pseudo-labels to overcome their inherent noise issues and enhance classification performance.

\subsection{Novelty of the Proposed PPLG:}
\textbf{Innovation in WSI Classification:}
WSI classification introduces distinct challenges, notably the reliance on bag-level labels due to the absence of instance-level annotations in a weakly supervised setting. Our PPLG approach overcomes this by generating high-quality pseudo-labels for instance-level training, crucial for accurate predictions at both the instance and bag levels. This represents the first integration of instance-level weakly supervised contrastive learning with prototype learning and an instance classifier in a cohesive training strategy, effectively using true negative instances from negative bags to guide learning.

\textbf{Innovation Among Related Works:}
Existing research, as outlined in Section \ref{PL}, utilizes prototype methods that typically extract patch features, develop prototypes through various clustering or similarity comparisons, and aggregate them using attention scores. However, these approaches often focus primarily on enhancing bag features and may lack precision in detailed instance classification. They also tend to rely heavily on attention scores, which can inaccurately identify challenging positive instances. Additionally, the integration of feature-level contrastive learning is frequently missing, impacting overall performance. In contrast, our PPLG strategy integrates instance contrastive learning and prototype learning with an instance classifier. This method effectively utilizes classifier guidance and incorporates true negative instances, significantly enhancing pseudo-label quality.

\textbf{Important Role in the Joint Training Strategy:}
The PPLG is integral to our joint training strategy, which includes IWSCL and the instance classifier. This holistic approach ensures synergistic contributions from each component, enhancing the overall model efficacy with the PPLG strategy playing a crucial role in refining the pseudo-labels for training the instance classifier.

\section{Conclusion}

In this paper, we propose INS, an instance-level MIL framework based on contrastive learning and prototype learning, which effectively addresses both instance and bag classification tasks. 
Guided by true negative instances, we propose a weakly-supervised contrastive learning method for effective instance-level feature representation under the MIL setting. 
In addition, we propose a prototype-based pseudo label generation method that generates high-quality pseudo labels for instances from positive bags. 
We further propose a joint training strategy for weakly-supervised contrastive learning, prototype learning, and instance classification. Extensive experiments on one synthetic 
dataset and five tasks on three real datasets demonstrate the strong performance of INS. With only slide labels, INS has the ability to accurately locate positive instances and 
has the potential to discover new knowledge or perform interpretability research on tough clinical tasks.

%\printbibliography
\bibliographystyle{IEEEtran}
\bibliography{INS_abbre}

% \vspace{-3.5em}
\begin{IEEEbiography}[{\includegraphics[width=1in,height=1.25in,clip,keepaspectratio]{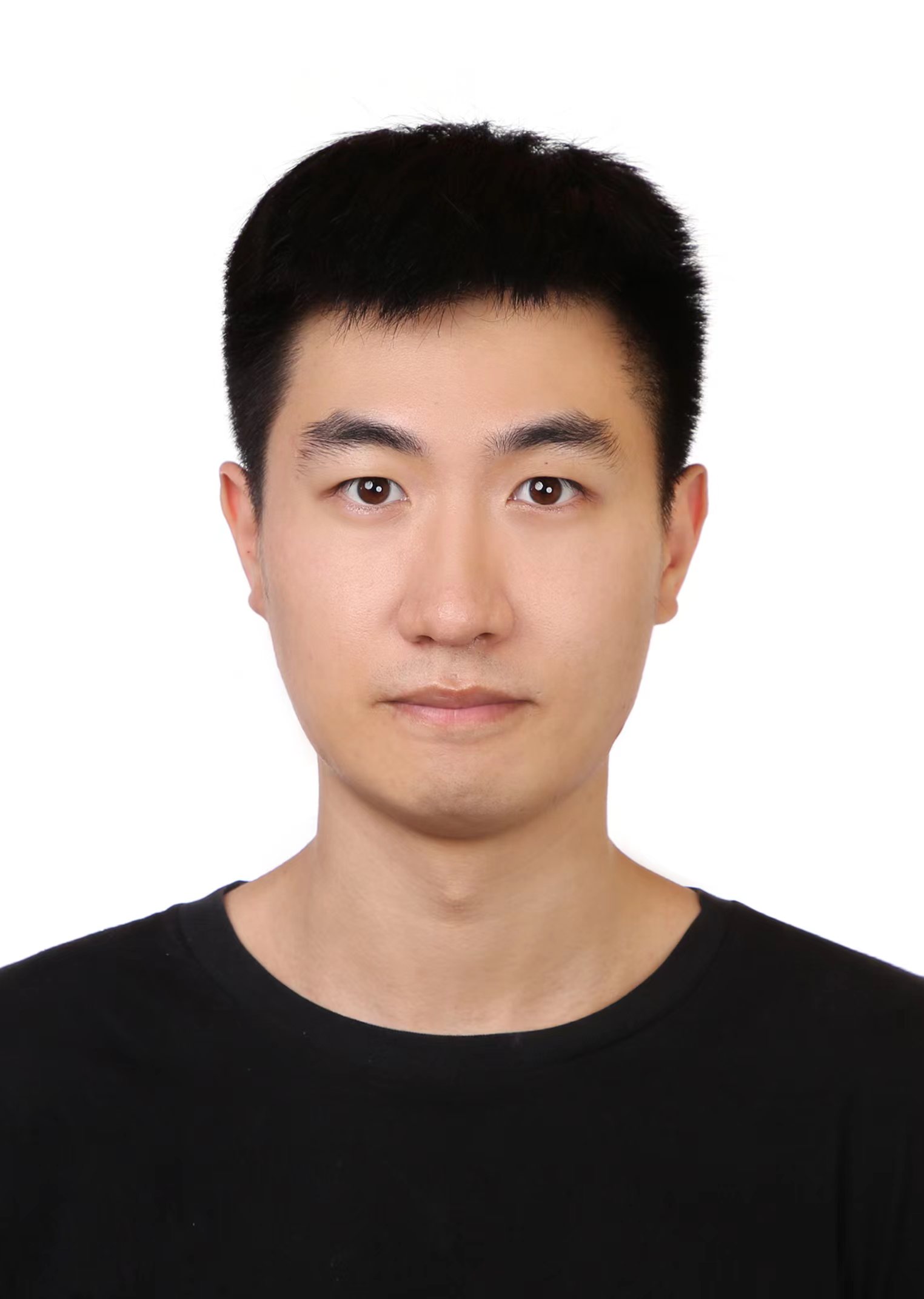}}]{Linhao Qu} is a Ph.D. student majoring in Biomedical Engineering at the School of Basic Medical Sciences, Fudan University. His research focuses on computational pathology, medical image processing, information fusion, and data mining. He has published over 20 papers in renowned international conferences and journals, appearing in prestigious international academic conferences such as NeurIPS, ICCV, CVPR, AAAI, MICCAI, and reputable journals like IEEE TMI, IEEE JBHI, Information Fusion, Modern Pathology and American Journal of Pathology.
\end{IEEEbiography}
  % \vspace{-3.5em}
\begin{IEEEbiography}[{\includegraphics[width=1in,height=1.25in,clip,keepaspectratio]{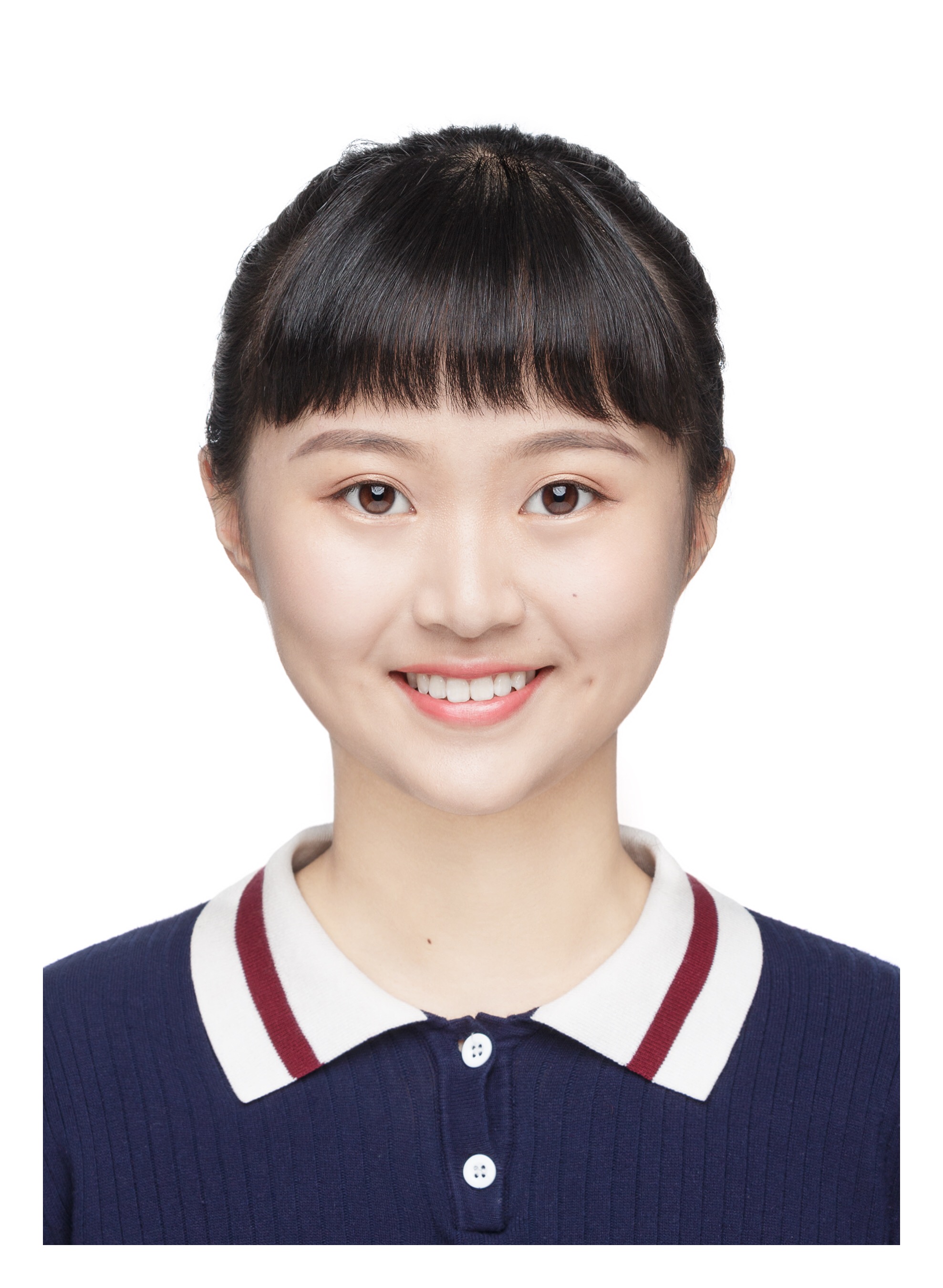}}]{Yingfan Ma} received the B.S. degree in computer science from the Harbin Institute of Technology(HIT) in 2022. She  is currently pursuing the Ph.D. degree in biomedical engineering from the Digital Medical Research Center, School of Basic Medical Sciences, Fudan University, Shanghai, China. Her research interests include computer vision and data mining.
\end{IEEEbiography}
  % \vspace{-3.5em}
  \begin{IEEEbiography}[{\includegraphics[width=1in,height=1.25in,clip,keepaspectratio]{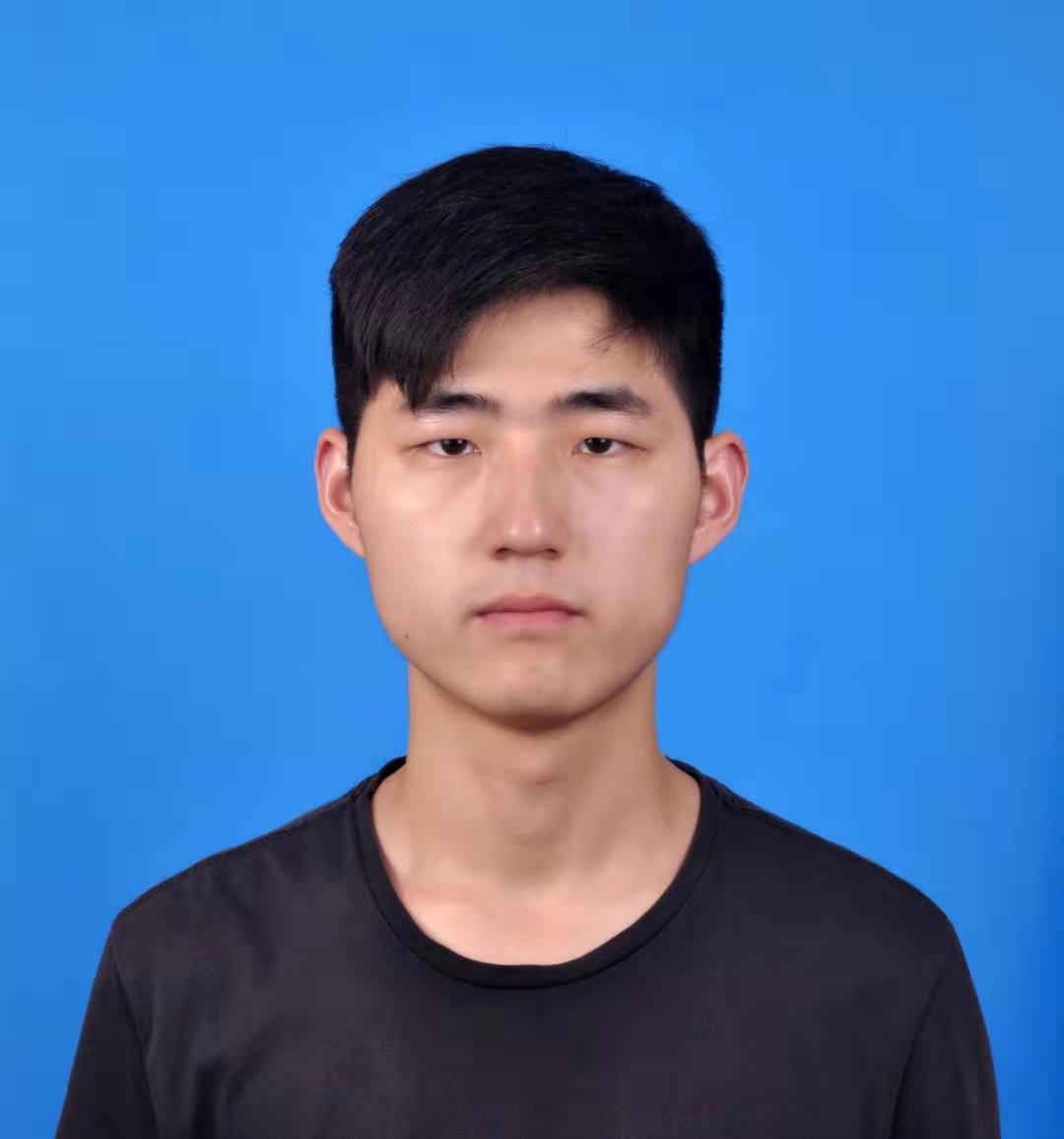}}]{Xiaoyuan Luo} received the B.S. degrees in automation from Wuhan University of Technology, Wuhan, China, in 2015. He is pursing doctoral degree in School of Basic Medical Science of Fudan University. His research interests include medical image processing and computer vision. He has published papers in top conferences and journals including NeurIPS, AAAI, MICCAI, IEEE JBHI.
\end{IEEEbiography}
  % \vspace{-3.5em}
\begin{IEEEbiography}[{\includegraphics[width=1in,height=1.25in,clip,keepaspectratio]{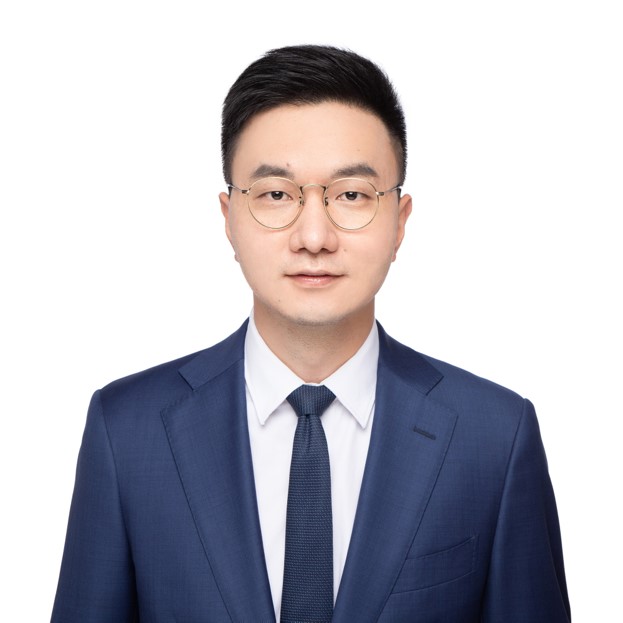}}]{Qinhao Guo} Guo is currently an attending physician in the Department of Gynecologic Oncology at the Affiliated Cancer Hospital of Fudan University. His main research interests include gynecologic oncology and computational pathology, and he has published more than ten papers in reputable and top journals.
  \end{IEEEbiography}
\begin{IEEEbiography}[{\includegraphics[width=1in,height=1.25in,clip,keepaspectratio]{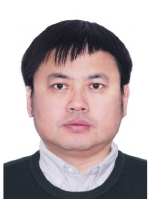}}]{Manning Wang} received the B.S. and M.S. degrees
    in power electronics and power transmission
    from Shanghai Jiaotong University, Shanghai,
    China, in 1999 and 2002, respectively. He
    received Ph.D. in biomedical engineering from
    Fudan University in 2011. He is currently a
    professor of biomedical engineering in School
    of Basic Medical Science of Fudan University.
    His research interests include medical image
    processing, image-guided intervention and computer
    vision.
\end{IEEEbiography}
  % \vspace{-3.5em}
\begin{IEEEbiography}[{\includegraphics[width=1in,height=1.25in,clip,keepaspectratio]{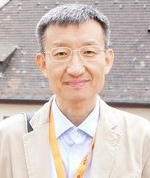}}]{Zhijian Song} received the Ph.D. degree in biomedical
    engineering from Xi’an Jiaotong University,
    Xi’an, China, in 1994. He is currently a Professor
    in School of Basic Medical Science of Fudan
    University, Shanghai, China, where he is also the
    Director of the Digital Medical Research Center
    and the Shanghai Key Laboratory of Medical
    Image Computing and Computer Assisted Intervention (MICCAI). His
    research interests include medical image processing, image-guided
    intervention, and the application of virtual and augmented reality technologies
    in medicine.
  \end{IEEEbiography}

\end{document}